\documentclass{article}
\usepackage{formatting/iclr2027_conference,times}

\let\cite\citep

\usepackage[utf8]{inputenc} 
\usepackage[T1]{fontenc}    
\usepackage{hyperref}       
\usepackage{url}            
\usepackage{booktabs}       
\usepackage{amsfonts}       
\usepackage{nicefrac}       
\usepackage{microtype}      
\usepackage{xcolor}         

\usepackage{amsmath, amssymb, amsthm}
\theoremstyle{definition}
\newtheorem{proposition}{Proposition}

\usepackage{formatting/packages}
\usepackage{formatting/hadi-colors}
\usepackage{formatting/hadi-macros-2025}
\usepackage{formatting/aliases}
\usepackage{formatting/formal-aliases}
\usepackage{subcaption}
\newcommand{\rc}[0]{Reasoning Compiler\xspace}

\newlength{\panellabelwidth}
\newcommand{\leftpanelfigure}[1]{%
  \begin{minipage}[t]{\panellabelwidth}
    \vspace{0pt}
    \raggedleft\textbf{(\thesubfigure)}
  \end{minipage}%
  \hspace{0.5em}%
  \begin{minipage}[t]{\dimexpr\linewidth-\panellabelwidth-0.5em\relax}
    \vspace{0pt}
    \includegraphics[width=\linewidth]{#1}
  \end{minipage}%
}

\title{\coltheading: More LLMs, Less Cost through \\ Collaborative Reasoning for Compiler \\ Optimizations}

\author{%
  Annabelle Sujun Tang, Christopher Priebe, Lianhui Qin, Hadi Esmaeilzadeh \\
  University of California San Diego \\
  \texttt{\{sujun,cpriebe,lianhui,hadi\}@ucsd.edu}
}

\iclrfinalcopy
\begin{document}

\maketitle

\begin{abstract}
LLM-guided compiler optimization has recently shown promise.
However, current LLM-guided optimizers query a single LLM at every search step, so their cost grows with both that model's price and the search budget.
A smaller model is cheaper but degrades the search: in our experiments, a recent single-LLM optimizer reaches 10.2\% (6.0\%) lower speedup on GPU (CPU) with gpt-5-mini than with GPT-5.2, which gives up the output quality that motivates using LLMs in the first place.
Agentic multi-LLM frameworks add controllers, concurrent calls, and shared memory, whose overhead can offset the savings of cheaper models.
This paper shows that the optimization search tree itself can coordinate heterogeneous LLMs: the tree already records which proposals improved the program, and we can augment the tree to record which model made them.
That evidence can then be used to decide which model should propose next.
To this end, we introduce \colt, in which each Monte Carlo tree search (MCTS) node records a program and the LLM that expands it, and the active LLM proposes both a compiler transformation and the LLM for the next step.
We also contribute an LLM-aware UCT that favors smaller models and a course-alteration rule that invokes the largest model only after repeated small-model regressions.
Against that single-LLM optimizer run with GPT-5.2 alone and the same number of search iterations and tree parameters, on an NVIDIA GPU (Intel CPU), \colt with eight LLMs improves the geometric-mean speedup on five kernels by 7.2\% (14.4\%) and the end-to-end Llama-3-8B speedup by 1.61$\times$ (1.41$\times$), while reducing kernel compilation time by 1.95$\times$ (1.74$\times$) and API cost by 4.47$\times$ (4.32$\times$), with GPT-5.2 making only 23.1\% (23.9\%) of LLM calls.
With the same eight models, random or round-robin model choice improves over the single-LLM baseline by only 2.5\% on CPU, showing that the gains come from letting the LLMs choose the next model rather than from mixing the pool at random.
\end{abstract}

\section{Introduction}
\label{sec:introduction}

The cost of model serving increasingly dominates modern AI systems, and the compiler determines how efficiently neural workloads run on target hardware through transformations such as scheduling, fusion, and layout changes~\cite{llm-serving-cost, dnn-compiler-survey:tpds:2020}.
Finding effective transformation sequences is difficult: the search space is combinatorial, hardware-dependent, and shaped by long-range interactions among transformations~\cite{tensor-comprehensions:arxiv:2018, tvm, ansor:osdi:2020, metaschedule:neurips:2022}.
LLMs are therefore entering the compiler optimization loop: LLM-guided optimizers condition each transformation proposal on the program, its transformation history, and observed outcomes, and they reach high-quality code in fewer samples than stochastic autotuners~\cite{llm-compiler-optimization:arxiv:2023, llm-compiler:cc:2025, compilerdream:kdd:2025, compiler-r1:neurips:2025, reasoning-compiler:neurips:2025}.

However, this quality comes at a price.
Existing LLM-guided optimizers query a single LLM at every search step~\cite{compiler-r1:neurips:2025, reasoning-compiler:neurips:2025}, so their cost scales with both that model's price and the search budget: a recent single-LLM optimizer---\rc---with \gptfivetwo spends 2262\,s and \$14.07 to tune a single Llama-4-Scout MLP kernel on CPU (Appendix~\ref{sec:runtime-cost}), and a full model contains many such kernels.
Using a smaller model alone does not solve this: in our experiments, \rc with \gptfivemini reaches 10.2\% (6.0\%) lower speedup on GPU (CPU) than with \gptfivetwo (\fref{fig:gpt5.2-speedup}), which gives up the output quality that motivates using LLMs at all.
This cost recurs for every model, hardware target, and update, whereas the speedup of the compiled code pays off at every inference.
As LLMs move into the compiler loop, their cost becomes a barrier alongside their quality.

Our goal in this work is to lower the cost of LLM-guided compiler optimization while retaining the output quality of a single large LLM, without fine-tuning, inter-model message passing, or agentic machinery.
Combining several LLMs is the natural direction, but LLM multi-agent systems coordinate through explicit interaction patterns, such as peer-to-peer exchange, coordinator-executor hierarchies, shared memory, debate, voting, or verifier-based aggregation~\cite{autogen:arxiv:2023, metagpt:iclr:2024, multi-agent-data-science:arxiv:2026, llmdebate:icml:2024, self-consistency:iclr:2023, moa:iclr:2025, multiagentverif:colm:2025}, and their concurrent calls, controllers, and memory state add overhead~\cite{should-we-be-going-mad:icml:2024, why-do-multi-agent-llm-systems-fail:neurips:2026} that can negate the savings of cheaper models.

\begin{figure*}[t]
\centering
\includegraphics[width=\linewidth]{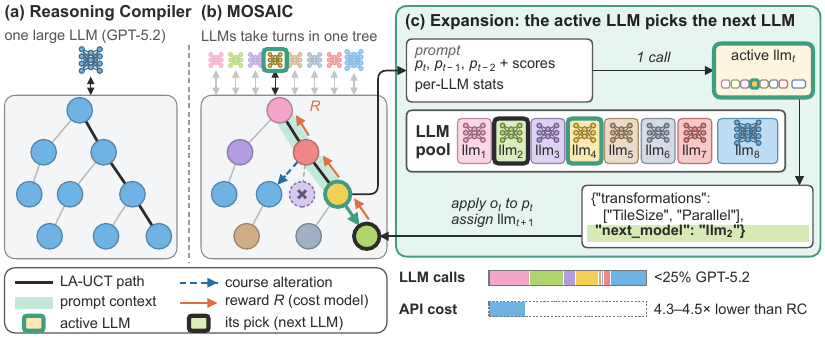}

\caption{\colt's LLMs collaborate only through one MCTS tree. (a) \rc uses one large LLM. (b) In \colt, small LLMs and \gptfivetwo take turns (color = expanding LLM, green = active): the active LLM reads nodes others wrote (green band), \mauct favors small LLMs, \gptfivetwo proposes a replacement child after two consecutive small-LLM regressions (dashed), and $R$ credits every LLM on the path. (c) One call to $\model_t$ returns $\langle\mutator_t,\model_{t+1}\rangle$, picking the next LLM (black) from the pool.}
\label{fig:overview}
\end{figure*}

Our key insight is that the optimization search tree already records which proposals improved the program, and can easily be augmented to record which model made them.
This evidence can therefore be used by an LLM to decide which model should propose next.
To this end, we introduce \colt, which makes model selection endogenous to one shared Monte Carlo tree search (MCTS)\nocite{mcts}: each search state pairs the current program with the active LLM, and each action jointly chooses a compiler transformation and the next LLM (\frefsub{fig:overview}{c}), so collaboration happens inside the search rather than through an external router or agent controller.

\AddToHookNext{cmd/@cflt/before}{%
  \edef\figureoneoldsep{\the\textfloatsep}%
  \setlength{\textfloatsep}{12pt}%
}
\AddToHookNext{cmd/@cflt/after}{%
  \setlength{\textfloatsep}{\figureoneoldsep}%
}

The shared MCTS tree is the central collaboration substrate in \colt (\frefsub{fig:overview}{b}): heterogeneous LLMs take turns expanding it, one call at a time, building on common transformation prefixes, and value estimates backpropagated through the same tree let signals found by one model influence later decisions made by others.
To reduce collaboration cost, we instantiate this idea with an LLM-aware Upper Confidence bounds applied to Trees (UCT)~\cite{uct:ecml:2006} variant that biases model choice toward smaller models while preserving the compiler performance objective. 
We further add a course-alteration mechanism that selectively invokes the largest model when persistent small model regressions would otherwise propagate degraded values.

We compare \colt against \rc~\cite{reasoning-compiler:neurips:2025}, the single-LLM MCTS optimizer that \colt extends (\frefsub{fig:overview}{a}), run with \gptfivetwo alone and the same number of search iterations and tree parameters.
On five kernels on an NVIDIA GPU (Intel CPU), \colteight improves the geometric-mean speedup by 7.2\% (14.4\%) while reducing compilation time by 1.95$\times$ (1.74$\times$) and API cost by 4.47$\times$ (4.32$\times$).
On CPU, choosing the next model at random from the same pool improves speedup by only 2.5\% and costs 1.27$\times$ more than \colt (\tref{tab:ablation}), so the gains come from letting the active LLM choose the next model.
Unlike LE-MCTS~\cite{park2025ensembling}, which picks models at random, and Multi-LLM AB-MCTS~\cite{inoue2025wider} and ShinkaEvolve~\cite{lange2025shinkaevolve}, which use bandits over per-model rewards, \colt lets the active LLM choose the next model from the program context and per-model statistics.

\section{\coltheading: Multi-LLM Tree Search for Compiler Optimization}
\label{sec:problem-formulation}
\label{sec:framework}
\colt makes one shared MCTS tree the coordination substrate for heterogeneous LLMs, without inter-model communication, concurrent LLM calls, or agentic controllers.

\subsection{Joint Search over Compiler Transformations and LLM Selection}
\label{sec:compiler-mdp}
\label{sec:joint-mdp}
\begingroup
\AddToHookNext{cmd/@makecol/before}{\raggedbottom}
\setlength{\abovedisplayskip}{4pt}
\setlength{\belowdisplayskip}{4pt}
\setlength{\abovedisplayshortskip}{0pt}
\setlength{\belowdisplayshortskip}{3pt}
We consider the task of optimizing an initial program $\program_1 \in \programset$ through a sequence of semantic-preserving compiler transformations~\cite{phase-ordering-problem:cf:2006}.
Each transformation $\mutator \in \mutatorset$ maps one program to another, and the objective is to identify a transformation sequence $\mutator_1,\ldots,\mutator_{T-1}$ that maximizes a performance metric $\objfunc : \programset \rightarrow \mathbb{R}_{\ge 0}$ on a target hardware platform:
\begin{equation*}
\program_{t+1} = \mutator_t(\program_t),
\qquad
\max_{\mutator_1,\ldots,\mutator_{T-1} \in \mutatorset}
\objfunc(\program_T).
\end{equation*}
Following standard formulations of compiler phase ordering, the underlying compiler environment can be modeled as a finite horizon Markov decision process (MDP)~\cite{phase-ordering-ml:oopsla:2012}: states are programs, actions are compiler transformations, and rewards are derived from the performance metric.
Transitions are stochastic in our implementation: an LLM names a transformation (e.g., TileSize), and the compiler samples its concrete parameters.

Let $\modelset = \{\model_1,\ldots,\model_N\}$ denote the available LLMs, which may differ in size, capability, and reliability.
To make collaboration possible and lightweight, we make the LLM choice part of the same decision process as the compiler optimization.
Thus, \colt augments the compiler state with the identity of the active LLM.
The joint state space is
\begin{equation*}
\mathcal{S} = \programset \times \modelset.
\end{equation*}
A state $\langle \program_t,\model_t\rangle$ contains the current program $\program_t$ and the LLM $\model_t$ responsible for proposing the next expansion.
A joint action is a pair
\begin{equation*}
\langle \mutator_t,\model_{t+1}\rangle \in \mutatorset \times \modelset,
\end{equation*}
where $\mutator_t$ is the compiler transformation applied to the current program and $\model_{t+1}$ is the model assigned to the resulting child state.
When the current state is $\langle \program_t,\model_t\rangle$, the active model $\model_t$ induces a context-dependent stochastic proposal distribution over such joint actions.
Applying a sampled joint action produces the next state
\begin{equation*}
\langle \program_{t+1}, \model_{t+1}\rangle
=
\langle \mutator_t(\program_t), \model_{t+1}\rangle.
\end{equation*}
For notation, we write $\mutator_t$ as a single transformation, although in the implementation an LLM response may contain a short sequence of valid compiler transformations that is composed and applied as one edge expansion.

\par\endgroup

\subsection{Shared-Tree MCTS with Endogenous Model Selection}
\label{sec:multi-llm-mcts}
\colt instantiates the joint search process above using MCTS.
\frefsub{fig:overview}{b} illustrates the optimization loop on the shared tree, and \frefsub{fig:overview}{c} one expansion.
Each tree node corresponds to a joint state $\langle \program,\model\rangle \in \programset \times \modelset$ and stores visit counts and accumulated rollout reward.
Each MCTS iteration proceeds through selection, expansion, rollout, and backpropagation.

During selection, the search descends the tree using the LLM-aware tree policy described in \sref{sec:ma-uct}.
The selected node determines both the current program and the active LLM.
During expansion, the active LLM is queried with the current optimization context and returns a joint proposal $\langle \mutator,\model'\rangle$, specifying the transformation to apply and the LLM to assign to the child node.
Applying $\mutator$ produces a new program state, yielding the child node $\langle \mutator(\program),\model'\rangle$.
After expansion, the search performs a short rollout consisting of randomly selected transformations.
The terminal program produced by the rollout is evaluated using a cost model, which estimates the objective without requiring direct execution on the target hardware at every search step.
The reward predicted by the cost model is then backpropagated along the selected path, updating visit counts and value estimates.
Since rewards depend only on resulting programs, credit assignment naturally spans both compiler transformations and the LLM selection decisions that led to them.

\subsection{LLM-Aware Tree Policy}
\label{sec:ma-uct}
In \colt, selecting a tree node determines both the program prefix to extend and the LLM that extends it, so the tree policy must weigh downstream program quality against LLM cost.
We therefore introduce \mauct, an LLM-aware variant of UCT~\cite{uct:ecml:2006} that scores each node by combining the compiler performance objective with a normalized preference for smaller LLMs.

As shown in Appendix~\ref{sec:mauct-proof}, at any fixed parent node treated as a stationary bandit, \mauct is equivalent to UCB1 applied to the transformed reward $(1-\lambda)R+\lambda\phi_{\mathrm{small}}(\model)$ and asymptotically concentrates visits on children maximizing the corresponding surrogate mean.
For a child node corresponding to program $\program_t$ and model $\model_t$, with parent $\langle \program_{t-1},\model_{t-1}\rangle$, we define
\begin{equation*}
\begin{aligned}
\text{\mauct}\left(\program_t, \model_t\right)
&=
(1-\lambda)\,
\frac{\mctscost\!\left(\program_t, \model_t\right)}
     {N\!\left(\program_t, \model_t\right)}
+
\lambda\,
\phi_{\mathrm{small}}\!\left(\model_t\right)
+
c
\sqrt{
\frac{
\ln N\!\left(\program_{t-1}, \model_{t-1}\right)
}{
N\!\left(\program_t, \model_t\right)
}
}.
\end{aligned}
\end{equation*}
Here, $N\!\left(\program_t,\model_t\right)$ is the visit count of the child node, $\mctscost\!\left(\program_t,\model_t\right)$ is its cumulative rollout reward, and $c$ controls exploration.
The first term estimates the downstream value of the program trajectory, while the second term favors smaller LLMs through
\begin{equation*}
\phi_{\mathrm{small}}\!\left(\model\right)
=
\frac{
\log{\parametercount{\model_{\max}}}
-
\log{\parametercount{\model}}
}{
\log{\parametercount{\model_{\max}}}
-
\log{\parametercount{\model_{\min}}}
+
\varepsilon
}
\in [0,1],
\end{equation*}
where $\parametercount{\model}$ denotes the parameter count of  $\model$, and $\model_{\min}$ and $\model_{\max}$ are the smallest and largest models in the candidate set.
Smaller models receive larger $\phi_{\mathrm{small}}$, and $\varepsilon$ prevents division by zero.
The size term is a cost prior known before search; observed reliability enters separately, through the per-model hit rates and error counts that the active LLM sees when it chooses the next model (\sref{sec:llm-model-selection}).
We use public parameter counts for open-weight models and assumed sizes for closed models, whose sizes are not disclosed (GPT-5.2: 300B; gpt-5-mini: 20B).
$\phi_{\mathrm{small}}$ depends on the assumed sizes only through log-ratios, and with the open-weight \llamathreeseventyb as the largest model, where only \gptfivemini's size is assumed, the results show the same trends (\sref{sec:results}).

The parameter $\lambda \in [0,1]$ controls the strength of the LLM size term.
When $\lambda=0$, selection reduces to reward-only UCT; as $\lambda$ increases, a child expanded by a smaller LLM is preferred when its expected downstream reward is competitive.
At $\lambda=0.5$, a child assigned to a larger model wins exploitation only if its mean reward exceeds the smaller model's by more than their gap in $\phi_{\mathrm{small}}$, so larger models are reached mainly through exploration, the active LLM's next-model choice, and course alteration.

\subsection{Contextual Prompting for Joint Transformation and LLM Selection}
\label{sec:llm-model-selection}
At each node expansion, the active LLM is queried for a joint proposal (\frefsub{fig:overview}{c}): a sequence of compiler transformations and the LLM to assign to the child node.
\niparagraph{Prompt context.}
For a selected leaf state $\langle \program_t,\model_t\rangle$, the prompt includes the current program $\program_t$ and, when available, its parent $\program_{t-1}$ and grandparent $\program_{t-2}$.
For these local program variants, \colt provides source code, transformation histories, and predicted performance scores.
The prompt also includes the valid transformations $\mutatorset$, the current leaf depth, and progress through the search budget.
The same prompt exposes the candidate LLMs ($\modelset$) with associated statistics collected during search.
For each LLM, \colt provides its size, invocation count, hit rate (the fraction of its expansions whose child scores higher than its parent under the cost model), and error count, where errors include invalid transformation names or invalid next model identifiers.
The prompt also includes local model context, namely the models used to expand the current node and its recent ancestors.

\niparagraph{Joint proposal.}
Given this combined context, the active LLM returns a structured proposal containing a transformation sequence and a single next model choice.
The LLM selection instruction is size-aware: prefer the smallest model likely to support continued progress, while allowing larger models when the local program context or prior statistics suggest that additional capacity may be useful.
The resulting child is then evaluated through the shared MCTS tree, so both the transformation proposal and the LLM recommendation receive credit only through their downstream effect on compiler reward.
Appendix~\ref{sec:prompt} provides a complete prompt template and an example output.

\subsection{Course Alteration via Large Model Intervention}
\label{sec:course-alteration}
Because \colt prefers smaller LLMs, one small model's errors can accumulate in the shared tree, so course alteration invokes the largest LLM when persistent small-model regressions are detected.
\niparagraph{Diagnosing persistent small-LLM regressions.}
A regression is detected when a child has a reward score worse than that of its parent.
A persistent small-model regression pattern is detected when two consecutive small-LLM expansions both produce regressions; expansions by the largest model neither count toward nor reset this streak.
When this pattern is detected, \colt asks the largest LLM for an alternative to the most recent regressive expansion, which replaces the regressive child unless it scores lower under the cost model, so a replaced child's degraded value estimate is never backpropagated through the tree.
%
%
%

\niparagraph{Large-model course alteration.}
\colt invokes the largest LLM from the same parent node using a dedicated course-alteration prompt (dashed arrow in \frefsub{fig:overview}{b}).
This prompt reuses the local program context, adds the failed small model proposal, and asks the largest LLM to revise the transformation sequence, the next LLM choice, or both.
This prompt is shorter and more targeted than a regular prompt, which lowers the cost of each high-capacity intervention.
An example course-alteration prompt is provided in Appendix~\ref{sec:prompt}.
The resulting child goes through the same rollout and backpropagation, after which the standard shared-tree search resumes.
Discarding regressive children biases small-model value estimates upward; we accept this as a heuristic safeguard and quantify its effect in \tref{tab:ablation}.

\section{Evaluation}
\label{sec:evaluation}
\subsection{Experimental Setup}
\label{sec:methodology}

\begin{figure*}[!t]
  \centering

  \begin{subfigure}{\textwidth}
    \phantomsubcaption
    \label{fig:gpu_speedup}
    \leftpanelfigure{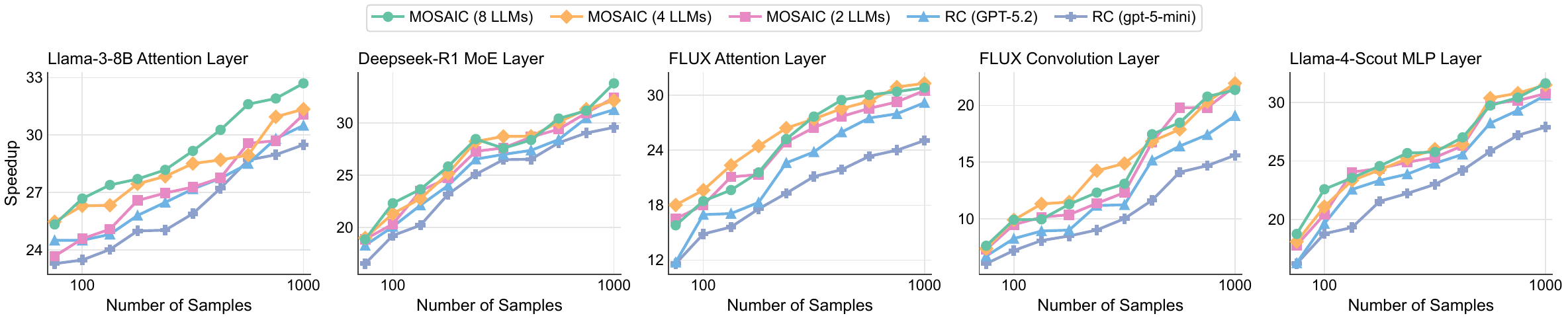}
  \end{subfigure}

  \begin{subfigure}{\textwidth}
    \phantomsubcaption
    \label{fig:gpt5.2-combined}
    \leftpanelfigure{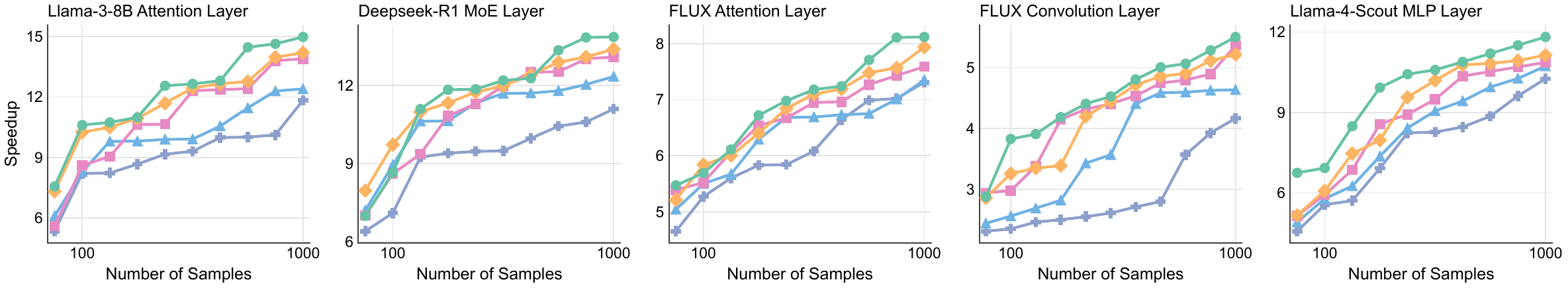}
  \end{subfigure}

  \caption{Speedup over unoptimized code as a function of the number of searched samples on \subref{fig:gpu_speedup}~GPU and \subref{fig:gpt5.2-combined}~CPU. RC is \rc~\cite{reasoning-compiler:neurips:2025} run with the named LLM alone; \colt uses \gptfivetwo as its largest model. Both panels share the legend of~\subref{fig:gpu_speedup}; Appendix~\ref{sec:final-speedups} lists the final values.}
  \label{fig:gpt5.2-speedup}
\end{figure*}

\begin{table*}[h]
    \centering
    \caption{Compilation time and API cost reduction of \colt relative to \rc run with the largest model alone. Column groups indicate the largest model. \textcolor{cerulean}{GPU}/CPU results are reported; \llamathreeseventyb results are CPU only.}
    \label{tab:litecoop_improvements}

    \renewcommand{\arraystretch}{1.08}
    \setlength{\tabcolsep}{4.0pt}
    \footnotesize

    \newcommand{\vcheader}[1]{\raisebox{-2.0ex}{\textbf{#1}}}

    \begin{tabular}{llcccccc}
        \toprule
        \multirow{2.5}{*}{\vcheader{Benchmark}} &
        \multirow{2.5}{*}{\vcheader{Metric}} &
        \multicolumn{3}{c}{%
            \makecell{\textbf{\gptfivetwo}}
        } &
        \multicolumn{3}{c}{%
            \makecell{\textbf{Llama-3.3-70B-Instruct}}
        } \\
    
        \cmidrule(lr){3-5} \cmidrule(l){6-8}
    
        & &
        \makecell{{\scriptsize\sffamily\color{coltcolor}\textbf{MOSAIC}}\\    {\scriptsize\sffamily\color{coltcolor}\textbf{(8 LLMs)}}} & 
        \makecell{{\scriptsize\sffamily\color{coltcolor}\textbf{MOSAIC}}\\    {\scriptsize\sffamily\color{coltcolor}\textbf{(4 LLMs)}}} & 
        \makecell{{\scriptsize\sffamily\color{coltcolor}\textbf{MOSAIC}}\\    {\scriptsize\sffamily\color{coltcolor}\textbf{(2 LLMs)}}} & 
        \makecell{{\scriptsize\sffamily\color{coltcolor}\textbf{MOSAIC}}\\    {\scriptsize\sffamily\color{coltcolor}\textbf{(8 LLMs)}}} & 
        \makecell{{\scriptsize\sffamily\color{coltcolor}\textbf{MOSAIC}}\\    {\scriptsize\sffamily\color{coltcolor}\textbf{(4 LLMs)}}} & 
        \makecell{{\scriptsize\sffamily\color{coltcolor}\textbf{MOSAIC}}\\    {\scriptsize\sffamily\color{coltcolor}\textbf{(2 LLMs)}}} \\
    
        \midrule
            
        \multirow{2}{*}{\makecell[l]{Llama-3-8B\\Attention Layer}} 
        & Comp. Time $\downarrow$ ($\times$)
        & \textcolor{cerulean}{1.85}/1.48 & \textcolor{cerulean}{1.46}/1.09 & \textcolor{cerulean}{1.01}/1.00
        & 1.93 & 1.68 & 1.31 \\
        
        & API Cost $\downarrow$ ($\times$)
        & \textcolor{cerulean}{4.35}/3.06 & \textcolor{cerulean}{3.30}/2.57 & \textcolor{cerulean}{1.87}/2.64 
        & 2.04 & 1.38 & 1.14 \\
        
        \midrule
        
        \multirow{2}{*}{\makecell[l]{DeepSeek-R1\\MoE Layer}} 
        & Comp. Time $\downarrow$ ($\times$)
        & \textcolor{cerulean}{1.79}/1.68 & \textcolor{cerulean}{1.20}/1.49 & \textcolor{cerulean}{1.05}/1.10 
        & 1.87 & 1.75 & 1.35 \\
        
        & API Cost $\downarrow$ ($\times$)
        & \textcolor{cerulean}{5.18}/4.18 & \textcolor{cerulean}{2.88}/3.21 & \textcolor{cerulean}{2.55}/2.42 
        & 2.35 & 1.59 & 1.15 \\
        
        \midrule
        
        \multirow{2}{*}{\makecell[l]{FLUX \\ Attention Layer}} 
        & Comp. Time $\downarrow$ ($\times$)
        & \textcolor{cerulean}{2.01}/2.03 & \textcolor{cerulean}{1.66}/1.18 & \textcolor{cerulean}{1.04}/1.10 
        & 1.84 & 1.62 & 1.35 \\
        
        & API Cost $\downarrow$ ($\times$)
        & \textcolor{cerulean}{5.37}/4.59 & \textcolor{cerulean}{3.22}/2.89 & \textcolor{cerulean}{1.87}/2.37 
        & 1.91 & 1.39 & 1.16 \\
        
        \midrule
        
        \multirow{2}{*}{\makecell[l]{FLUX \\ Convolution Layer}} 
        & Comp. Time $\downarrow$ ($\times$)
        & \textcolor{cerulean}{2.09}/1.73 & \textcolor{cerulean}{1.47}/1.22 & \textcolor{cerulean}{1.06}/1.02 
        & 2.23 & 1.53 & 1.31 \\
        
        & API Cost $\downarrow$ ($\times$)
        & \textcolor{cerulean}{3.76}/5.87 & \textcolor{cerulean}{3.18}/2.96 & \textcolor{cerulean}{2.78}/2.17 
        & 2.30 & 1.34 & 1.16 \\
        
        \midrule
        
        \multirow{2}{*}{\makecell[l]{Llama-4-Scout\\MLP Layer}} 
        & Comp. Time $\downarrow$ ($\times$)
        & \textcolor{cerulean}{2.03}/1.85 & \textcolor{cerulean}{1.75}/1.10 & \textcolor{cerulean}{1.08}/1.04 
        & 2.32 & 1.72 & 1.25 \\
        
        & API Cost $\downarrow$ ($\times$)
        & \textcolor{cerulean}{3.91}/4.38 & \textcolor{cerulean}{3.10}/2.56 & \textcolor{cerulean}{2.63}/2.23 
        & 2.67 & 1.42 & 1.12 \\
        
        \bottomrule
    \end{tabular}
\end{table*}

\begin{table*}[h]
    \centering
    \caption{Invocation rates (\%) of each model, averaged across the five benchmarks, for 2-, 4-, and 8-LLM configurations of \colt on \textcolor{cerulean}{GPU}/CPU (\llamathreeseventyb: CPU only). C.A.\ denotes course-alteration calls.} 
    \label{tab:model_invocation}

    \renewcommand{\arraystretch}{1.08}
    \setlength{\tabcolsep}{1.6pt}
    \footnotesize
    
    \begin{tabular}{lccc|lccc}
        \toprule
        \multicolumn{4}{c|}{\textbf{Largest Model: \gptfivetwo \textcolor{cerulean}{GPU}/CPU}} & 
        \multicolumn{4}{c}{\textbf{Largest Model: \llamathreeseventyb (CPU)}} \\
        
        \cmidrule(r){1-4} \cmidrule(l){5-8}
        
        \makecell[l]{Model} & 
        \makecell{{\scriptsize\sffamily\color{coltcolor}\textbf{MOSAIC}}\\{\scriptsize\sffamily\color{coltcolor}\textbf{(8 LLMs)}}} & 
        \makecell{{\scriptsize\sffamily\color{coltcolor}\textbf{MOSAIC}}\\{\scriptsize\sffamily\color{coltcolor}\textbf{(4 LLMs)}}} & 
        \makecell{{\scriptsize\sffamily\color{coltcolor}\textbf{MOSAIC}}\\{\scriptsize\sffamily\color{coltcolor}\textbf{(2 LLMs)}}} & 
        
        \makecell[l]{Model} & 
        \makecell{{\scriptsize\sffamily\color{coltcolor}\textbf{MOSAIC}}\\{\scriptsize\sffamily\color{coltcolor}\textbf{(8 LLMs)}}} & 
        \makecell{{\scriptsize\sffamily\color{coltcolor}\textbf{MOSAIC}}\\{\scriptsize\sffamily\color{coltcolor}\textbf{(4 LLMs)}}} & 
        \makecell{{\scriptsize\sffamily\color{coltcolor}\textbf{MOSAIC}}\\{\scriptsize\sffamily\color{coltcolor}\textbf{(2 LLMs)}}} \\
        
        \midrule
        
        \gptfivetwo (Regular) & \textcolor{cerulean}{9.8}/10.7 & \textcolor{cerulean}{16.1}/17.8 & \textcolor{cerulean}{26.2}/24.5 & 
        Llama-3.3-70B-Inst (Reg.) & 18.6 & 30.5 & 45.5 \\

        \makecell[l]{\gptfivetwo (C.A.)} & \textcolor{cerulean}{13.3}/13.2 & \textcolor{cerulean}{12.2}/12.0 & \textcolor{cerulean}{10.6}/10.5 & 
        \makecell[l]{Llama-3.3-70B-Inst (C.A.)} & 12.1 & 10.1 & 7.9 \\

        \gptfivetwo (Total) & \textcolor{cerulean}{23.1}/23.9 & \textcolor{cerulean}{28.3}/29.8 & \textcolor{cerulean}{36.7}/35.1 & 
        Llama-3.3-70B-Inst (Total) & 30.7 & 40.6 & 53.4 \\

        \midrule
        
        \gptfivemini & \textcolor{cerulean}{1.8}/7.8 & \textcolor{cerulean}{33.3}/36.5 & \textcolor{cerulean}{63.3}/64.9 & 
        \gptfivemini & 3.0 & 27.9 & 46.6 \\
        
        DS-R1-Qwen-7B & \textcolor{cerulean}{25.4}/12.6 & --- & --- & 
        DS-R1-Qwen-7B & 8.9 & --- & --- \\
        
        Llama-3.1-8B-Inst & \textcolor{cerulean}{7.4}/11.4 & \textcolor{cerulean}{29.2}/15.1 & --- & 
        Llama-3.1-8B-Inst & 15.6 & 23.1 & --- \\
        
        Qwen3-8B & \textcolor{cerulean}{21.9}/35.2 & --- & --- & 
        Qwen3-8B & 22.0 & --- & --- \\

        Qwen3-14B & \textcolor{cerulean}{14.8}/4.7 & --- & --- & 
        Qwen3-14B & 15.2 & --- & --- \\

        Devstral-Small-2505 & \textcolor{cerulean}{1.1}/2.4 & --- & --- & 
        Devstral-Small-2505 & 3.5 & --- & --- \\

        DS-R1-Qwen-32B & \textcolor{cerulean}{4.5}/2.0 & \textcolor{cerulean}{9.2}/18.6 & --- & 
        DS-R1-Qwen-32B & 1.0 & 8.5 & --- \\
        
        \bottomrule
    \end{tabular}
\end{table*}

\begin{table*}[ht]
    \centering
    \caption{End-to-end \llamathreeb: final speedup improvement, compilation time reduction, and API cost reduction of \colt over \rc run with the largest model alone. \textcolor{cerulean}{GPU}/CPU results are reported; \llamathreeseventyb results are CPU only.}
    \label{tab:e2e_efficiency}
    
    \setlength{\tabcolsep}{1.9pt}
    \footnotesize
    
    \begin{tabular}{lccc|ccc}
        \toprule
        \multicolumn{4}{c|}{\textbf{Largest Model: \gptfivetwo}} & 
        \multicolumn{3}{c}{\textbf{Largest Model: \llamathreeseventyb}} \\
        
        \cmidrule(r){1-4} \cmidrule(l){5-7}
        
        \makecell[l]{ MOSAIC \\ Configuration} & 
        \makecell{Speedup over \\ RC $\uparrow$ ($\times$)} & 
        \makecell{Comp. \\ Time  $\downarrow$ ($\times$)} & 
        \makecell{API Cost \\ $\downarrow$ ($\times$)} & 
        
        \makecell{Speedup over \\ RC $\uparrow$ ($\times$)} & 
        \makecell{Comp. \\ Time $\downarrow$ ($\times$)} & 
        \makecell{API Cost \\ $\downarrow$ ($\times$)} \\
        
        \midrule
        
        \smallcoltconfig{8} & \textcolor{cerulean}{1.61}/1.41 & \textcolor{cerulean}{1.52}/1.71 & \textcolor{cerulean}{3.94}/4.44 & 
        1.34 & 1.45 & 1.96 \\
        
        \smallcoltconfig{4} & \textcolor{cerulean}{1.86}/1.19 & \textcolor{cerulean}{1.32}/1.27 & \textcolor{cerulean}{2.81}/2.79 & 
        1.27 & 1.39 & 1.55 \\
        
        \smallcoltconfig{2} & \textcolor{cerulean}{1.11}/1.28 & \textcolor{cerulean}{1.22}/1.10 & \textcolor{cerulean}{2.19}/2.21 & 
        1.23 & 1.28 & 1.36 \\
        
        \bottomrule
    \end{tabular}
\end{table*}

\begin{table*}[t]
    \centering
    \caption{Where the gains come from, on CPU at 1000 samples. All \colt variants use the same eight-LLM pool and the same shared tree; only the listed component changes, and indented rows modify \colteight. RC is \rc. The last two columns give compilation time and API cost relative to \colt (higher is more expensive).}
    \label{tab:ablation}
    \renewcommand{\arraystretch}{1.05}
    \setlength{\tabcolsep}{2.6pt}
    \footnotesize
    \begin{tabular}{lccccccccc}
        \toprule
        Variant & \makecell{L3\\Attn.} & \makecell{MoE} & \makecell{FLUX\\Attn.} & \makecell{FLUX\\Conv.} & \makecell{L4\\MLP} & \makecell{Geo-\\mean} & \makecell{vs.\\RC} & Time & Cost \\
        \midrule
        RC (\gptfivetwo only) & 12.40 & 12.35 & 7.35 & 4.64 & 10.73 & 8.91 & --- & 1.74$\times$ & 4.32$\times$ \\
        Random next LLM & 12.82 & 12.54 & 7.37 & 4.93 & 10.83 & 9.12 & +2.5\% & 1.37$\times$ & 1.27$\times$ \\
        Round-robin next LLM & 12.96 & 12.69 & 7.42 & 4.82 & 10.80 & 9.13 & +2.5\% & 1.38$\times$ & 1.26$\times$ \\
        \textbf{\colteight} & \textbf{14.98} & \textbf{13.87} & \textbf{8.12} & \textbf{5.51} & \textbf{11.82} & \textbf{10.19} & \textbf{+14.4\%} & 1.00$\times$ & 1.00$\times$ \\
        \quad w/o size term ($\lambda{=}0$) & 14.62 & 13.65 & 7.85 & 5.44 & 11.58 & 9.97 & +12.0\% & --- & --- \\
        \quad w/o course alteration & 14.05 & 12.91 & 7.61 & 5.33 & 11.36 & 9.65 & +8.3\% & --- & --- \\
        \quad alteration after every regression & 15.26 & 14.24 & 8.18 & 5.53 & 11.92 & 10.32 & +15.9\% & 1.38$\times$ & 1.58$\times$ \\
        \bottomrule
    \end{tabular}
\end{table*}

\niparagraph{Benchmarks and platforms.}
We evaluate five kernels from production-scale neural networks: (1) a self-attention layer from \llamathreeb~\cite{llama3:arxiv:2024}, (2) a mixture-of-experts (MoE) layer from \deepseekrone~\cite{deepseek:arxiv:2025}, (3) a self-attention layer and (4) a convolution layer from \flux~\cite{flux:gitrepo:2024}, and (5) an MLP layer from \llamafourscout~\cite{llama4:huggingface:2025}.
We also compile end-to-end \llamathreeb.
We report results on an NVIDIA 2080 Ti GPU and an Intel Core i9 CPU.

\niparagraph{Baselines.}
Our single-LLM baseline is \rc~\cite{reasoning-compiler:neurips:2025}, the LLM-guided MCTS compiler optimizer that \colt extends.
We run \rc in our environment with \gptfivetwo or \gptfivemini as its only model and with the same search budget, cost model, exploration constant, branching factor, and final-program selection as \colt.
We do not compare against non-LLM autotuners: our question is how to lower the cost of LLM-guided optimization, whose sample-efficiency advantage over TVM MetaSchedule on these kernels was established on CPUs by \citet{reasoning-compiler:neurips:2025} in a separate campaign (Appendix~\ref{sec:final-speedups}).

\niparagraph{\colt configurations.}
We evaluate three model sets of increasing size: (1) \colttwo comprises \gptfivetwo and \gptfivemini; (2) \coltfour adds DeepSeek-R1-Distill-Qwen-32B~\cite{deepseek:arxiv:2025} and Llama-3.1-8B-Instruct~\cite{llama3:arxiv:2024}; and (3) \colteight further adds DeepSeek-R1-Distill-Qwen-7B~\cite{deepseek:arxiv:2025}, Qwen3-8B, Qwen3-14B~\cite{qwen3technicalreport}, and Devstral-Small-2505~\cite{devstral-small-2505}.
Going from four to eight LLMs adds only open-weight models of 7B--24B parameters, none larger than the 32B model or \gptfivetwo already in the pool, so the gains of \colteight do not come from adding a larger model.

\niparagraph{Implementation.}
We implement \colt in TVM's MetaSchedule~\cite{tvm, metaschedule:neurips:2022} (Apache TVM v0.20.0~\cite{tvm:v0200:2025}) by replacing its search routine, and we access models through the OpenAI and Nscale APIs; the code is in the Supplementary Material (Appendix~\ref{sec:code}).
\mauct uses $\lambda = 0.5$ (Appendix~\ref{sec:lambda}), exploration constant $c = \sqrt{2}$, and branching factor $B = 2$, following prior work~\cite{branching1:cg:2007, ucb:ml:2002, reasoning-compiler:neurips:2025}.
We set $\lambda$ and the course-alteration threshold on the CPU kernels (Appendices~\ref{sec:lambda} and~\ref{sec:course-alteration-ablation}) and use them unchanged on GPU and with \llamathreeseventyb.
Each experiment is repeated 10 times, and we report the mean.

\niparagraph{Samples, rewards, and measurement.}
One sample is one MCTS iteration: one node expansion by one regular LLM call, followed by a rollout, so every 1000-sample run makes exactly 1000 regular LLM calls; course-alteration calls are additional and are included in \colt's compilation time, API cost, and call counts.
Rollouts are scored by TVM's default XGBoost cost model~\cite{tvm, xgboost:kdd:2016}, which MetaSchedule trains online on candidates measured on the target hardware; \rc uses the same model, so neither method has a cost-model advantage.
\colt itself needs only a scalar prediction per program, which any regression-based cost model provides and which an order-preserving map (e.g., clipping or min--max scaling) brings into $[0,1]$, as is standard in cost-model-guided autotuning~\cite{autotvm:neurips:2020, ansor:osdi:2020, metaschedule:neurips:2022}.
During search, the reward is the cost model's prediction clipped below at 0; the model regresses each measured candidate's throughput relative to the best measured candidate of the same workload, a target in $(0,1]$, so rewards lie approximately in $[0,1]$ but are not bounded above.
All speedups in figures and tables, however, are latencies measured on the target hardware: speedup is the latency of the unoptimized IRModule divided by the lowest measured latency among the programs found within the given number of samples, averaged over the 10 runs.

\niparagraph{Correctness.}
The LLMs never emit code.
They name MetaSchedule schedule transformations (tile sizes, compute location, parallelization, and unrolling), whose parameters MetaSchedule samples and which change loop structure and memory placement but not the computation, and TVM rejects any proposal that is invalid for the program.
Every candidate is therefore semantically equivalent to the input program by construction, up to floating-point reassociation from reordered reductions, as in MetaSchedule itself~\cite{metaschedule:neurips:2022}.

\niparagraph{Cost accounting.}
Compilation time is the wall-clock time of the entire search, including LLM latency, rollouts, and hardware measurement.
API cost sums, over all calls, the input and output tokens times each model's list price from its provider (OpenAI for \gptfivetwo and \gptfivemini; Nscale for open-weight models).
Both measures depend on the serving provider: wall-clock time includes queueing under variable load and network latency, and list prices differ across providers.
They are nevertheless what users of hosted models pay, so we report them (Appendix~\ref{sec:runtime-cost}) along with per-model call counts (\tref{tab:model_invocation}), which do not depend on the provider.

\subsection{Experimental Results}
\label{sec:results}
\niparagraph{\colt retains the quality of a large LLM at a fraction of its cost.}
A small model alone degrades the search: at 1000 samples, \rc with \gptfivemini trails \rc with \gptfivetwo by 10.2\% (6.0\%) on GPU (CPU) (\fref{fig:gpt5.2-speedup}).
\colt recovers and exceeds the quality of the large model.
On GPU (\fref{fig:gpu_speedup}), \colteight reaches a geometric-mean speedup of 29.7$\times$ and improves over \rc with \gptfivetwo on all five kernels, by 7.2\% on geometric mean and up to 11.9\% on \fluxconvbench (19.1$\times$ to 21.4$\times$).
On CPU (\fref{fig:gpt5.2-combined}), \colteight reaches 10.2$\times$ and improves over \rc with \gptfivetwo on all five kernels, by 14.4\% on geometric mean and up to 20.8\% on \llamabench (12.4$\times$ to 15.0$\times$).

These gains come at lower cost: relative to \rc with \gptfivetwo, \colteight reduces compilation time by 1.95$\times$ (1.74$\times$) and API cost by 4.47$\times$ (4.32$\times$) on GPU (CPU) (\tref{tab:litecoop_improvements}), since small models perform most expansions (\tref{tab:model_invocation}).

\niparagraph{On CPU, the LLM-chosen next model drives the speedup gains.}
\tref{tab:ablation} isolates each component of \colt on CPU with the same eight-LLM pool and shared tree (full results in Appendices~\ref{sec:lambda}--\ref{sec:model-selection-ablation}).
Random or round-robin next-LLM selection, i.e., a static mixture of the pool, improves over \rc by only 2.5\%, and it costs 1.37$\times$ (1.38$\times$) more compilation time and 1.27$\times$ (1.26$\times$) more API cost than \colt.
Letting the active LLM choose the next model raises the gain to 14.4\%, and removing the size term ($\lambda{=}0$, reward-only UCT) or course alteration lowers it to 12.0\% or 8.3\%, respectively.
The two mechanisms are complementary: the LLM-chosen next model adds 11.9 points over random choice, and course alteration adds 6.1 points.
Invoking course alteration after every regression gains only 1.5 more points but costs 1.38$\times$ more time and 1.58$\times$ more API cost.
Every tested $\lambda \in \{0, 0.25, 0.5, 0.75, 1\}$ beats \rc on every kernel (+9.3\% to +14.4\%; Appendix~\ref{sec:lambda}), and $\lambda{=}0.5$ beats both reward-only ($\lambda{=}0$) and size-only ($\lambda{=}1$) node selection, so both terms of \mauct contribute.
We conclude that the benefit of \colt comes from treating model choice as part of the joint MCTS action rather than from mixing the pool at random.

\niparagraph{Scaling the LLM set improves the cost-quality tradeoff.}
Growing the pool from two to four to eight LLMs raises the geometric-mean final GPU speedup from 29.0$\times$ to 29.4$\times$ to 29.7$\times$, and the compilation-time (API-cost) reduction from 1.05$\times$ (2.31$\times$) to 1.50$\times$ (3.13$\times$) to 1.95$\times$ (4.47$\times$), since larger pools shift more expansions to small models.
Eight LLMs are not the best everywhere, however: on GPU, \coltfour beats \colteight on \fluxattbench and \fluxconvbench at 1000 samples and on end-to-end \llamathreeb (1.86$\times$ vs.\ 1.61$\times$ over \rc; \tref{tab:e2e_efficiency}).
Cost, in contrast, falls at each step from two to four to eight LLMs on both platforms (on CPU, 1.05$\times$, 1.21$\times$, and 1.74$\times$ lower compilation time).
\colteight therefore offers the best cost-quality tradeoff rather than the highest speedup on every workload.

\niparagraph{Small models perform most expansions.}
As \tref{tab:model_invocation} shows, the total \gptfivetwo invocation rate falls from 36.7\% (35.1\%) with \colttwo to 23.1\% (23.9\%) with \colteight on GPU (CPU), and regular \gptfivetwo calls fall to only 9.8\% (10.7\%); the rest are course alterations, which use a shorter prompt.

\niparagraph{End-to-end compilation.}
On end-to-end \llamathreeb (\tref{tab:e2e_efficiency}), \colt improves final speedup over \rc with \gptfivetwo by 1.11--1.86$\times$ on GPU and 1.19--1.41$\times$ on CPU, while reducing compilation time by 1.10--1.71$\times$ and API cost by 2.19--4.44$\times$.

\niparagraph{\colt's gains hold with an open-weight largest LLM.}
With the open-weight \llamathreeseventyb as the largest model (CPU; Appendix~\ref{sec:llama70}), \colteight improves over \rc with \llamathreeseventyb by 4.8--27.5\% at 1000 samples on every kernel.
It also reduces compilation time by 1.84--2.32$\times$ and API cost by 1.91--2.67$\times$, and it invokes the largest model for only 30.7\% of calls.

\subsection{Limitations}
\label{sec:limitations}
We evaluate five kernels, spanning attention, mixture-of-experts, convolution, and MLP operators from four model families, and one end-to-end model on one consumer GPU and one CPU within TVM; extending \colt to data-center GPUs and other compilers is future work.
The mechanism needs only a scalar reward per node, but we evaluate it only on compiler optimization.
We report means over 10 runs; \colteight beats \rc with the same largest model on all 15 kernel settings (Appendix~\ref{sec:final-speedups}), as does every $\lambda$ and course-alteration setting on CPU (Appendices~\ref{sec:lambda}--\ref{sec:course-alteration-ablation}).
The size preference uses parameter count as a proxy for cost, and closed-model sizes are assumed (\sref{sec:ma-uct}).
We do not evaluate each small model alone, bandit-based model selection~\cite{inoue2025wider, lange2025shinkaevolve, symphony:neurips:2025}, cheap-first cascades~\cite{automix:neurips:2024}, separate per-model trees, or debate- and orchestrator-style multi-agent baselines (see also Appendix~\ref{app:limitations}).

\section{Related Work}
\label{sec:related-work}
\niparagraph{Multi-LLM tree search.}
LE-MCTS~\cite{park2025ensembling} ensembles LLMs inside MCTS for mathematical reasoning, choosing a model at random per expansion and scoring steps with a pre-trained process reward model.
Multi-LLM AB-MCTS~\cite{inoue2025wider} and ShinkaEvolve~\cite{lange2025shinkaevolve} instead select the LLM that expands a node or mutates a program with a bandit (Thompson sampling or UCB1) over per-model rewards, SYMPHONY~\cite{symphony:neurips:2025} assembles heterogeneous LLM agents in MCTS planning, and AlphaEvolve~\cite{alphaevolve:arxiv:2025} mixes a fast and a strong LLM for code evolution.
\colt differs in who chooses and why: the active LLM itself proposes the next model from the program context and per-model statistics, the tree policy prefers smaller models, and the largest model is escalated to after regressions, so the pool is used to cut cost rather than only to raise quality.
Our random next-LLM ablation (\tref{tab:ablation}) uses LE-MCTS's uniform model choice and improves over \rc by only 2.5\%.
\niparagraph{Adaptive LLM use under cost constraints.}
Confidence-based cascades and cost-aware routers balance LLM quality and cost per query~\cite{frugalgpt:arxiv:2023, hybridllm:iclr:2024, automix:neurips:2024, routellm:arxiv:2024, routing-and-cascading:arxiv:2025, universal-model-routing:arxiv:2025, router-r1:neurips:2025, slm-mux:iclr:2026}.
\colt instead assigns credit to model choices through their downstream effect on search reward, without a fixed or trained router.
\niparagraph{Collaborative reasoning across LLMs.}
Multi-agent LLM systems coordinate through debate, critique, or message passing~\cite{llmdebate:icml:2024, multiagentverif:colm:2025, moa:iclr:2025, autogen:arxiv:2023, metagpt:iclr:2024, multi-agent-data-science:arxiv:2026, self-consistency:iclr:2023}.
\colt coordinates models only through the states and rewards of a shared MCTS tree, with one LLM call per expansion.
\niparagraph{ML for compiler optimization.}
Compiler optimization has used both search and learning~\cite{ml-in-compilers-survey:ieee:2018}, including MCTS without LLMs~\cite{protuner:arxiv:2020, koo2021customized}, LLMs~\cite{llm-compiler-optimization:arxiv:2023, priority-sampling-llm-compiler:arxiv:2024, llm-compiler:cc:2025, compilerdream:kdd:2025, decos:ics:2025, compiler-r1:neurips:2025}, language models trained to generate tensor schedules~\cite{tlm:osdi:2024}, and their combination in \rc~\cite{reasoning-compiler:neurips:2025}.

\section{Conclusion}
We introduced \colt, which turns a shared MCTS tree into the collaboration mechanism for heterogeneous LLMs, so that small models drive most decisions and \colt finds faster code than a single large LLM at a fraction of its cost.
LLM-guided optimizers need not scale by calling ever larger models, but by letting many inexpensive models share one search and calling the largest model mostly to correct course.

\newpage

\bibliographystyle{formatting/iclr2027_conference}
\bibliography{bib/full-ref}

\newpage
\appendix

\section{Local Equivalence and Concentration of LA-UCT}
\label{sec:mauct-proof}

Proposition~\ref{prop:lauct} formalizes the local behavior of the LA-UCT rule introduced in Section~\ref{sec:ma-uct}.

\begin{proposition}[Local equivalence and concentration of LA-UCT]
\label{prop:lauct}
Fix a parent node \(u\) with a fixed, finite set of children \(v_1,\ldots,v_K\), where \(v_j\) is expanded by model \(\model_t^{(j)}\), and assume:
(A1) the rewards \(R_{j,s}\) observed at \(v_j\) are i.i.d.\ across visits with mean \(\mu_j\);
(A2) \(R_{j,s} \in [0,1]\);
(A3) the maximizer \(j_\lambda = \argmax_j \widetilde{\mu}_j\) of \(\widetilde{\mu}_j = (1-\lambda)\mu_j + \lambda\phi_{\mathrm{small}}(\model_t^{(j)})\) is unique; and
(A4) \(c = \sqrt{2}\).
Then selecting children of \(u\) by LA-UCT is UCB1 applied to the transformed rewards \(Y_{j,s} = (1-\lambda)R_{j,s} + \lambda\phi_{\mathrm{small}}(\model_t^{(j)})\), and every \(j \neq j_\lambda\) satisfies
\(\mathbb{E}[N(v_j)] \le 8\ln N(u)/(\Delta_j^{\lambda})^2 + 1 + \pi^2/3\), where \(\Delta_j^{\lambda} = \widetilde{\mu}_{j_\lambda} - \widetilde{\mu}_j\).
\end{proposition}

\noindent\textit{Scope.} In \colt, (A1) and (A2) hold only approximately: rewards come from a cost model that MetaSchedule retrains online, subtrees grow and course alteration can replace children, so child rewards are not stationary, and predictions are clipped below at 0 but not above (Section~\ref{sec:methodology}). With branching factor \(B=2\) and 1000 iterations, most parents also receive few visits. The proposition therefore describes the intended local behavior of the selection rule rather than a guarantee for our runs.

\begin{proof}
Let
\(
u = \langle\program_{t-1}, \model_{t-1}\rangle
\)
be a parent node. Suppose that \(u\) has finitely many children
\[
v_j = \bigl\langle\program_t^{(j)}, \model_t^{(j)}\bigr\rangle,
\qquad j = 1,\ldots,K.
\]
Let \(R_{j,s}\) be the reward observed the \(s\)-th time node \(v_j\) is selected; by (A2), \(R_{j,s} \in [0,1]\). Any regression-based cost model's output can be brought into this range by an order-preserving map, such as clipping or min--max scaling. In our implementation, the predictions of TVM's XGBoost model approximate normalized throughput in \((0,1]\) and are clipped below at 0 (Section~\ref{sec:methodology}).

The LA-UCT score of child \(v_j\) is
\[
\operatorname{LA\text{-}UCT}(v_j)
=
(1-\lambda)\frac{W(v_j)}{N(v_j)}
+
\lambda \phi_{\mathrm{small}}\bigl(\model_t^{(j)}\bigr)
+
c\sqrt{\frac{\ln N(u)}{N(v_j)}}.
\]

Define the transformed reward
\[
Y_{j,s}
=
(1-\lambda)R_{j,s}
+
\lambda \phi_{\mathrm{small}}\bigl(\model_t^{(j)}\bigr).
\]
Since
\(
R_{j,s} \in [0,1]
\; \text{and} \;
\phi_{\mathrm{small}}\bigl(\model_t^{(j)}\bigr) \in [0,1],
\)
we have
\(
Y_{j,s} \in [0,1].
\)

Then
\[
\begin{aligned}
\overline{Y}_j
&=
\frac{\sum_{s=1}^{N(v_j)} Y_{j,s}}{N(v_j)} \\[0.5em]
&=
\frac{
\sum_{s=1}^{N(v_j)}
\left(
(1-\lambda)R_{j,s}
+
\lambda \phi_{\mathrm{small}}\bigl(\model_t^{(j)}\bigr)
\right)
}{N(v_j)} \\[0.5em]
&=
\frac{
(1-\lambda)W(v_j)
+
N(v_j)\lambda \phi_{\mathrm{small}}\bigl(\model_t^{(j)}\bigr)
}{N(v_j)} \\[0.5em]
&=
(1-\lambda)\frac{W(v_j)}{N(v_j)}
+
\lambda \phi_{\mathrm{small}}\bigl(\model_t^{(j)}\bigr).
\end{aligned}
\]

Therefore,
\[
\operatorname{LA\text{-}UCT}(v_j)
=
\overline{Y}_j
+
c\sqrt{\frac{\ln N(u)}{N(v_j)}}.
\]
Thus, at a fixed parent node, LA-UCT is UCT applied to the transformed reward
\[
Y_{j,s}
=
(1-\lambda)R_{j,s}
+
\lambda \phi_{\mathrm{small}}\bigl(\model_t^{(j)}\bigr).
\]

Define the transformed mean
\[
\widetilde{\mu}_j
=
\mathbb{E}[Y_{j,s}]
=
(1-\lambda)\mathbb{E}[R_{j,s}]
+
\lambda \phi_{\mathrm{small}}\bigl(\model_t^{(j)}\bigr).
\]
Let
\(
\mu_j = \mathbb{E}[R_{j,s}].
\)
Then
\(
\widetilde{\mu}_j
=
\mathbb{E}[Y_{j,s}]
=
(1-\lambda)\mu_j
+
\lambda \phi_{\mathrm{small}}\bigl(\model_t^{(j)}\bigr).
\) By (A3), the maximizer
\(
j_{\lambda}
=
\argmax_j \widetilde{\mu}_j
\)
is unique.

For each suboptimal child \(j \neq j_{\lambda}\), define the surrogate gap
\[
\Delta_j^{\lambda}
=
\widetilde{\mu}_{j_{\lambda}}
-
\widetilde{\mu}_j
>
0.
\]

Since the transformed rewards satisfy
\[
Y_{j,s} \in [0,1],
\]
we can apply Theorem 1 of \citet{ucb:ml:2002} to the bandit with arm means \(\widetilde{\mu}_j\), optimal arm \(j_{\lambda}\), and pull counts \(N(v_j)\). By (A1), (A2), and (A4), the theorem applies to UCB1 with index \(\overline{Y}_j + \sqrt{2\ln N(u)/N(v_j)}\) and gives, for each suboptimal arm \(j\),
\[
\mathbb{E}[N(v_j)]
\leq
\frac{8 \ln N(u)}{\bigl(\Delta_j^{\lambda}\bigr)^2}
+
1 + \frac{\pi^2}{3}.
\]

Therefore, LA-UCT concentrates its visits on the child maximizing
\[
\widetilde{\mu}_j
=
(1-\lambda)\mu_j
+
\lambda \phi_{\mathrm{small}}\bigl(\model_t^{(j)}\bigr).
\]

At any fixed parent node, LA-UCT is equivalent to UCB1 applied to a transformed reward that combines expected downstream program reward with the normalized preference for smaller models. Therefore, the policy allocates a vanishing fraction of visits to children that are suboptimal with respect to this surrogate mean, and asymptotically concentrates on children that best balance downstream reward and model size preference. In this sense, the proposition formalizes the intended size-aware behavior of LA-UCT: smaller LLMs are favored when they are sufficient, while larger LLMs remain preferred whenever they provide sufficiently greater expected downstream value.
\end{proof}

\newpage

\clearpage
\section{Prompt}
\label{sec:prompt}

\begin{promptbox}[Regular Model Invocation Prompt]

You are an AI scheduling assistant to help with a Monte Carlo Tree Search (MCTS) to find an optimal program in the search space starting from an unoptimized program.

In this MCTS, the current program is the leaf we are expanding, while immediate parent and grandparent refer to the ancestors in the tree.

Each program has:
\begin{itemize}
    \item a piece of code
    \item a transformation history sequence
    \item a predicted performance score
\end{itemize}

You are given:
\begin{itemize}
    \item Code of the current program
    \item Historical performance info of current/parent/grandparent (Pieces of code, transformation history sequences, predicted scores)
    \item A list of possible transformations that can be applied next
    \item Search context: leaf depth and trials progress
    \item Global per-model stats:
    \begin{itemize}
        \item hit\_rate: fraction of calls where score(child) $>$ score(parent)
        \item number of times the model is invoked
        \item number of errors the model has made in past invocations (+1 for invalid transformation, +1 for invalid next\_model name)
        \item parameter counts 
    \end{itemize}
    \item Local model context:
    \begin{itemize}
        \item model used to expand current/parent/grandparent
    \end{itemize}
\end{itemize}

\textbf{Task:}
\begin{enumerate}
    \item Compare code/transformation history/predicted performance scores to infer what changes might improve performance.
    \item Propose a sequence of transformations from the provided list. You may repeat a transformation to explore different decisions.
    \item Choose exactly one model from the provided model list as the next model to expand the child. Use the smallest model that could give best results. Prefer models with fewer errors.
\end{enumerate}

Output a single valid JSON object in the EXACT format:
\begin{lstlisting}[frame=none, numbers=none]
{
  "transformations": ["Fullname1", "Fullname2", "..."],
  "next_model": "..."
}
\end{lstlisting}

\textbf{Historical Performance Info (Leaf, Parent, Grandparent)}

\textbf{Current Program:}

Code:
\begin{lstlisting}[language=python, frame=none, numbers=none]
@T.prim_func
def main(A, B, C):
    C_global = T.alloc_buffer((1, 16, 4096))
    for fused in T.parallel(2):
        for b_1, i_1, j_1 in T.grid(1, 2, 64):
            ...
            for k_0, ..., k_1, ... in T.grid(512, ..., 8, ...):
                for j_vec in T.vectorized(64):
                    with T.block("matmul_update"):
                        ...
                        C_global[...] = C_global[...] + A[...] * B[...]
        for ax0, ax1, ax2 in T.grid(1, 8, 4096):
            C[...] = C_global[...]
\end{lstlisting}

Transformation history:
\begin{lstlisting}[language=python, frame=none, numbers=none]
sch.sample_perfect_tile(loop=j, decision=[1, 64, 1, 64])
sch.vectorize(...)
sch.cache_write(..., storage_scope="global")
sch.decompose_reduction(...)
...
\end{lstlisting}

Predicted score: 0.0739

\textbf{Immediate Parent Schedule:}

Code:
\begin{lstlisting}[language=python, frame=none, numbers=none]
@T.prim_func
def main(A, B, C):
    C_global = T.alloc_buffer((1, 16, 4096))
    for fused in T.parallel(4):
        for b_1, i_1, j_1 in T.grid(1, 2, 32):
            ...
            for k_0, ..., k_1, ... in T.grid(512, ..., 8, ...):
                for j_vec in T.vectorized(32):
                    with T.block("matmul_update"):
                        ...
                        C_global[...] = C_global[...] + A[...] * B[...]
        for ax0, ax1, ax2 in T.grid(1, 8, 2048):
            C[...] = C_global[...]
\end{lstlisting}

Transformation history:
\begin{lstlisting}[language=python, frame=none, numbers=none]
sch.sample_perfect_tile(loop=j, decision=[2, 32, 2, 32])
sch.vectorize(...)
sch.cache_write(..., storage_scope="global")
sch.decompose_reduction(...)
...
\end{lstlisting}

Predicted score: 0.136

\textbf{Available Transformations}
\begin{lstlisting}[frame=none, numbers=none]
[
  "ComputeLocation",
  "Parallel",
  "Unroll",
  "TileSize"
]
\end{lstlisting}

\textbf{Search Context}

Leaf depth: 3 \\
Trials progress: 10 / 300

\textbf{Global Per-Model Stats}

Model gpt-5-mini: params=20.0B, regular\_calls=12, regular\_hit\_rate=0.364, errors=0 \\
Model gpt-5.2: params=300.0B, regular\_calls=11, regular\_hit\_rate=0.5, course\_alteration\_calls=0, course\_alteration\_hit\_rate=0, errors=0

\textbf{Local Model Context}

Model used to expand the current node: gpt-5.2\\
Model used to expand the parent node: gpt-5.2 \\
Model used to expand the grandparent node: N/A

\end{promptbox}

\begin{promptbox}[Model Answer]
{
  "transformations": [
    "TileSize",
    "TileSize",
    "ComputeLocation",
    "Parallel",
    "Unroll"
  ],
  "next\_model": "gpt-5-mini"
}

\end{promptbox}

\begin{promptbox}[Course Alteration Prompt]

You are the largest model invoked for course alteration in a Monte Carlo Tree Search (MCTS) for compiler optimization.
A smaller model has proposed a sequence of transformations and a next model for expanding the child node.
This proposal triggered course alteration because the predicted score of the resulting child is lower than the predicted score of the current program.

In this MCTS, the current program is the leaf being expanded, while immediate parent and grandparent refer to ancestors in the tree.
Each program has:
\begin{itemize}
    \item a piece of code
    \item a transformation history sequence
    \item a predicted performance score
\end{itemize}

You are given:
\begin{itemize}
    \item Code of the current program
    \item Code of the parent program
    \item A list of possible transformations that can be applied next
    \item The smaller model's proposed transformations and proposed next model
    \item The predicted score of the current program and the predicted score after applying the smaller model's proposal
    \item Search context: leaf depth and trials progress
    \item Global per-model stats:
    \begin{itemize}
        \item hit\_rate: fraction of calls where score(child) $>$ score(parent)
        \item number of times the model is invoked
        \item number of errors the model has made in past invocations (+1 for invalid transformation, +1 for invalid next\_model name)
        \item parameter counts
    \end{itemize}
    \item Local model context:
    \begin{itemize}
        \item model used to expand current/parent/grandparent
    \end{itemize}
\end{itemize}

\textbf{Task:}
\begin{enumerate}
    \item Modify the smaller model's proposal by changing the transformation sequence, the next model, or both. Compare code/predicted performance scores to infer what changes might improve performance.
    \item Propose a sequence of transformations from the provided list. You may repeat a transformation to explore different decisions.
    \item Choose exactly one model from the provided model list as the next model to expand the child. Use the smallest model that could give best results. Prefer models with fewer errors.
\end{enumerate}

Output a single valid JSON object in the EXACT format:
\begin{lstlisting}[frame=none, numbers=none]
{
  "transformations": ["Fullname1", "Fullname2", "..."],
  "next_model": "..."
}
\end{lstlisting}

\textbf{Historical Performance Info (Leaf, Parent, Grandparent)}

\textbf{Current Program:}

Code:
\begin{lstlisting}[language=python, frame=none, numbers=none]
@T.prim_func
def main(A, B, C):
    C_global = T.alloc_buffer((1, 16, 4096))
    for fused in T.parallel(2):
        for b_1, i_1, j_1 in T.grid(1, 4, 1):
            ...
            for k_0, ..., k_1, ... in T.grid(80, ..., 64, ...):
                for j_vec in T.vectorized(32):
                    with T.block("matmul_update"):
                        ...
                        C_global[...] = C_global[...] + A[...] * B[...]
        for ax0, ax1, ax2 in T.grid(1, 16, 2048):
            C[...] = C_global[...]
\end{lstlisting}

Predicted score: 0.460

\textbf{Immediate Parent Program:}

Code:
\begin{lstlisting}[language=python, frame=none, numbers=none]
@T.prim_func
def main(A, B, C):
    C_global = T.alloc_buffer((1, 16, 4096))
    for fused in T.parallel(64):
        for b_1, i_1, j_1 in T.grid(1, 8, 1):
            ...
            for k_0, ..., k_1, ... in T.grid(80, ..., 64, ...):
                for j_vec in T.vectorized(32):
                    with T.block("matmul_update"):
                        ...
                        C_global[...] = C_global[...] + A[...] * B[...]
        for ax0, ax1, ax2 in T.grid(1, 16, 256):
            C[...] = C_global[...]
\end{lstlisting}

Predicted score: 0.572

\textbf{Smaller Model Proposal Triggering Course Alteration}

Smaller model name: gpt-5-mini

Proposed transformations:
\begin{lstlisting}[frame=none, numbers=none]
["TileSize", "Parallel", "Unroll"]
\end{lstlisting}

Proposed next model: gpt-5.2

Predicted current score: 0.460

Predicted child score from smaller model proposal: 0.028

\textbf{Search Context}

Leaf depth: 9 \\
Trials progress: 179 / 300

\textbf{Global Per-Model Stats}

Model gpt-5-mini: params=20.0B, regular\_calls=425, regular\_hit\_rate=0.494, errors=0

Model gpt-5.2: params=300.0B, regular\_calls=232, regular\_hit\_rate=0.513, course\_alteration\_calls=73, course\_alteration\_hit\_rate=0.282, errors=0

\textbf{Local Model Context}

Model used to expand the current node: gpt-5-mini \\
Model used to expand the parent node: gpt-5-mini \\
Model used to expand the grandparent node: gpt-5.2

\end{promptbox}

\begin{promptbox}[Course Alteration Model Answer]
{
  "transformations": ["TileSize", "ComputeLocation"],
  "next\_model": "gpt-5-mini"
}
\end{promptbox}
\clearpage

\section{Code Repository}

The code implementation of \colt is available at \url{https://github.com/he-actlab/MOSAIC}.
\label{sec:code}

\newpage
\section{Per-Kernel Final Speedups}
\label{sec:final-speedups}

\tref{tab:final-speedups} lists the speedups at 1000 samples plotted in \fref{fig:gpt5.2-speedup} and \fref{fig:llama70-combined}.
\citet{reasoning-compiler:neurips:2025} compare \rc with TVM MetaSchedule on these CPU kernels in a separate experimental campaign; absolute speedups differ between the two campaigns, so that comparison indicates relative sample efficiency rather than absolute speedups in our setting.

\begin{table*}[h]
\centering
\caption{Speedup over unoptimized code at 1000 samples. RC is \rc run with the named LLM alone. CPU (70B) uses \llamathreeseventyb as the largest model.}
\label{tab:final-speedups}
\footnotesize
\setlength{\tabcolsep}{3.5pt}
\begin{tabular}{llcccccc}
\toprule
Platform & Method & \makecell{Llama-3\\Attn.} & \makecell{DS-R1\\MoE} & \makecell{FLUX\\Attn.} & \makecell{FLUX\\Conv.} & \makecell{L4-Scout\\MLP} & Geomean \\
\midrule
\multirow{5}{*}{GPU} & RC (\gptfivemini) & 29.49 & 29.56 & 25.05 & 15.60 & 27.90 & 24.86 \\
 & RC (\gptfivetwo) & 30.50 & 31.25 & 29.17 & 19.09 & 30.62 & 27.68 \\
 & \colttwo & 31.07 & 32.40 & 30.50 & 21.68 & 30.76 & 28.99 \\
 & \coltfour & 31.35 & 32.13 & 31.28 & 21.95 & 31.51 & 29.35 \\
 & \textbf{\colteight} & \textbf{32.70} & \textbf{33.77} & \textbf{30.80} & \textbf{21.37} & \textbf{31.64} & \textbf{29.67} \\
\midrule
\multirow{5}{*}{CPU} & RC (\gptfivemini) & 11.84 & 11.10 & 7.31 & 4.17 & 10.26 & 8.37 \\
 & RC (\gptfivetwo) & 12.40 & 12.35 & 7.35 & 4.64 & 10.73 & 8.91 \\
 & \colttwo & 13.90 & 13.10 & 7.59 & 5.37 & 10.87 & 9.58 \\
 & \coltfour & 14.20 & 13.40 & 7.94 & 5.22 & 11.14 & 9.74 \\
 & \textbf{\colteight} & \textbf{14.98} & \textbf{13.87} & \textbf{8.12} & \textbf{5.51} & \textbf{11.82} & \textbf{10.19} \\
\midrule
\multirow{5}{*}{\makecell[l]{CPU\\(70B)}} & RC (\gptfivemini) & 11.84 & 11.10 & 7.31 & 4.17 & 10.26 & 8.37 \\
 & RC (Llama-3.3-70B) & 12.50 & 10.82 & 7.42 & 3.82 & 10.80 & 8.38 \\
 & \colttwo & 12.95 & 11.49 & 7.53 & 4.32 & 10.97 & 8.81 \\
 & \coltfour & 12.73 & 11.91 & 7.72 & 4.38 & 11.72 & 9.03 \\
 & \textbf{\colteight} & \textbf{13.10} & \textbf{12.50} & \textbf{7.89} & \textbf{4.87} & \textbf{11.91} & \textbf{9.44} \\
\bottomrule
\end{tabular}
\end{table*}

\section{\llamathreeseventyb as the Largest Model}
\label{sec:llama70}

\begin{figure*}[h]
  \centering
  \includegraphics[width=\textwidth]{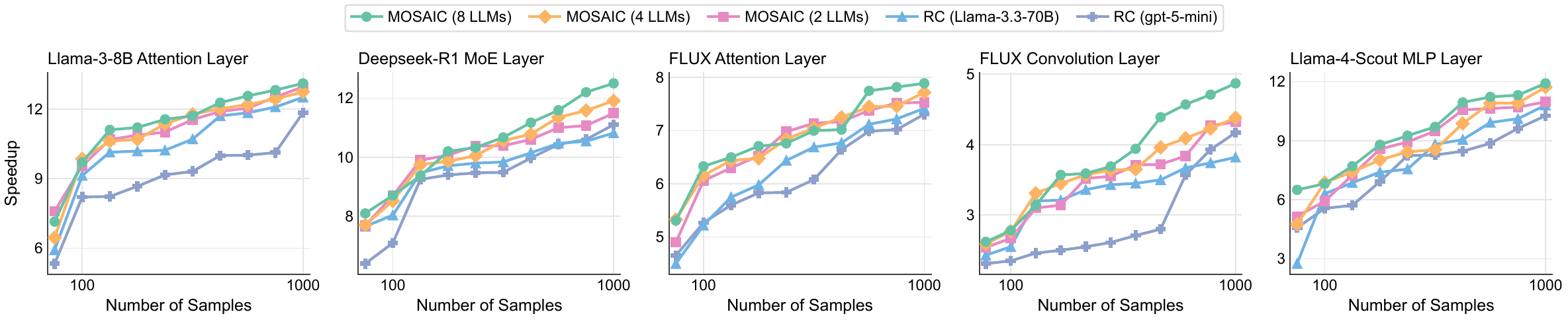}
  \caption{CPU speedup over unoptimized code as a function of the number of searched samples, using \llamathreeseventyb as the largest model. RC is \rc run with the named LLM alone.}
  \label{fig:llama70-combined}
\end{figure*}

\fref{fig:llama70-combined} shows CPU speedup when \llamathreeseventyb~\cite{llama3:arxiv:2024} replaces \gptfivetwo as the largest model, so that every model except \gptfivemini is open-weight with a public parameter count.
\colteight achieves the best final speedup on every kernel, improves over \rc with \llamathreeseventyb by 4.8--27.5\% at 1000 samples, and also gains early (e.g., 6.3$\times$ vs.\ 5.2$\times$ on \fluxattbench at 100 samples).
\rc with \gptfivemini and \rc with \llamathreeseventyb reach nearly the same geometric mean (8.37$\times$ vs.\ 8.38$\times$; \tref{tab:final-speedups}), and \colteight improves over both, to 9.44$\times$.

\clearpage
\section{Absolute Compilation Time and API Cost}
\label{sec:runtime-cost}

\begin{table*}[ht]
\centering
\caption{Absolute compilation time and API cost on CPU for \rc with \gptfivetwo and for \colt with 2, 4, and 8 LLMs.}
\label{tab:runtime-api-cost}
\resizebox{\textwidth}{!}{%
\begin{tabular}{@{}llcccc@{}}
\toprule

\multirow{5}{*}{Llama-3-8B Attention Layer}
 & Setting & Runtime (sec) & Runtime Imp. & API Cost & API Cost Imp. \\
\cmidrule{2-6}
 & \rc (GPT-5.2) & 2236 & --- & \$13.49 & --- \\
 & 2 Models & 2227 & 1.00 & \$5.11 & 2.64 \\
 & 4 Models & 2051 & 1.09 & \$5.25 & 2.57 \\
 & 8 Models & 1514 & 1.48 & \$4.41 & 3.06 \\
\midrule

\multirow{5}{*}{DeepSeek-R1 MoE Layer}
 & Setting & Runtime (sec) & Runtime Imp. & API Cost & API Cost Imp. \\
\cmidrule{2-6}
 & \rc (GPT-5.2) & 2493 & --- & \$13.75 & --- \\
 & 2 Models & 2270 & 1.10 & \$5.68 & 2.42 \\
 & 4 Models & 1675 & 1.49 & \$4.28 & 3.21 \\
 & 8 Models & 1488 & 1.68 & \$3.29 & 4.18 \\
\midrule

\multirow{5}{*}{FLUX Attention Layer}
 & Setting & Runtime (sec) & Runtime Imp. & API Cost & API Cost Imp. \\
\cmidrule{2-6}
 & \rc (GPT-5.2) & 2494 & --- & \$13.78 & --- \\
 & 2 Models & 2265 & 1.10 & \$5.82 & 2.37 \\
 & 4 Models & 2112 & 1.18 & \$4.77 & 2.89 \\
 & 8 Models & 1227 & 2.03 & \$3.00 & 4.59 \\
\midrule

\multirow{5}{*}{FLUX Convolution Layer}
 & Setting & Runtime (sec) & Runtime Imp. & API Cost & API Cost Imp. \\
\cmidrule{2-6}
 & \rc (GPT-5.2) & 2393 & --- & \$16.73 & --- \\
 & 2 Models & 2342 & 1.02 & \$7.71 & 2.17 \\
 & 4 Models & 1968 & 1.22 & \$5.65 & 2.96 \\
 & 8 Models & 1382 & 1.73 & \$2.85 & 5.87 \\
\midrule

\multirow{5}{*}{Llama-4-Scout MLP Layer}
 & Setting & Runtime (sec) & Runtime Imp. & API Cost & API Cost Imp. \\
\cmidrule{2-6}
 & \rc (GPT-5.2) & 2262 & --- & \$14.07 & --- \\
 & 2 Models & 2166 & 1.04 & \$6.32 & 2.23 \\
 & 4 Models & 2049 & 1.10 & \$5.49 & 2.56 \\
 & 8 Models & 1223 & 1.85 & \$3.21 & 4.38 \\
\bottomrule
\end{tabular}%
}
\end{table*}
\clearpage
\section{Ablations of Lambda Parameters}
\label{sec:lambda}

As shown in Table \ref{tab:lambda-ablation-speedups}, at 1000 searched samples $\lambda = 0.5$ gives the highest final speedup on every benchmark (geometric mean $10.19\times$), and every tested value improves over \rc. $\lambda = 0.5$ provides the best empirical balance between the two components of the LA-UCT surrogate reward: the downstream program reward term and the normalized model size preference term. With this balance, LA-UCT can preserve the compiler performance objective while still encouraging efficient use of smaller models when they are sufficient, as shown in Table \ref{tab:lambda-ablation-call-percentages}, allowing the search to allocate model choices more effectively and achieve the best overall speedup.

\begin{table*}[ht]
\centering
\caption{Speedup over unoptimized code across varying numbers of samples for different choices of \(\lambda\) values.}
\label{tab:lambda-ablation-speedups}

\begingroup
\renewcommand{\arraystretch}{0.95}

\resizebox{\textwidth}{!}{%
\begin{tabular}{@{}llcccccc@{}}
\toprule
 Benchmark & $\lambda$ Value / Number of Samples & 50 & 100 & 250 & 500 & 750 & 1000 \\
\midrule

\multirow{5}{*}{Llama-3-8B Attention Layer}
 & 0.00 & 7.36 & 10.22 & 10.93 & 12.27 & 12.50 & 14.62 \\
 & 0.25 & 7.49 & 10.49 & 10.82 & 12.51 & 12.70 & 14.85 \\
 & 0.50 & 7.56 & 10.61 & 10.99 & 12.65 & 12.81 & 14.98 \\
 & 0.75 & 7.30 & 10.19 & 10.61 & 12.16 & 12.40 & 14.52 \\
 & 1.00 & 7.09 & 9.87 & 10.78 & 11.77 & 12.09 & 14.15 \\
\midrule

\multirow{5}{*}{DeepSeek-R1 MoE Layer}
 & 0.00 & 7.22 & 9.01 & 11.64 & 12.05 & 12.37 & 13.65 \\
 & 0.25 & 7.08 & 8.82 & 11.71 & 12.12 & 12.34 & 13.80 \\
 & 0.50 & 6.99 & 8.66 & 11.84 & 12.20 & 12.28 & 13.87 \\
 & 0.75 & 6.97 & 8.62 & 11.54 & 11.99 & 12.24 & 13.56 \\
 & 1.00 & 6.95 & 8.44 & 11.36 & 11.87 & 12.26 & 13.44 \\
\midrule

\multirow{5}{*}{FLUX Attention Layer}
 & 0.00 & 5.26 & 5.64 & 6.43 & 6.99 & 7.07 & 7.85 \\
 & 0.25 & 5.35 & 5.58 & 6.58 & 7.09 & 7.18 & 8.02 \\
 & 0.50 & 5.47 & 5.69 & 6.72 & 7.18 & 7.24 & 8.12 \\
 & 0.75 & 5.33 & 5.62 & 6.55 & 7.06 & 7.15 & 7.96 \\
 & 1.00 & 5.19 & 5.55 & 6.40 & 6.95 & 7.03 & 7.74 \\
\midrule

\multirow{5}{*}{FLUX Convolution Layer}
 & 0.00 & 2.84 & 3.76 & 4.13 & 4.51 & 4.93 & 5.44 \\
 & 0.25 & 2.87 & 3.80 & 4.16 & 4.48 & 4.98 & 5.48 \\
 & 0.50 & 2.88 & 3.83 & 4.19 & 4.53 & 5.01 & 5.51 \\
 & 0.75 & 2.82 & 3.74 & 4.02 & 4.42 & 4.88 & 5.38 \\
 & 1.00 & 2.77 & 3.67 & 4.10 & 4.29 & 4.76 & 5.26 \\
\midrule

\multirow{5}{*}{Llama-4-Scout MLP Layer}
 & 0.00 & 6.43 & 6.78 & 9.62 & 10.31 & 10.80 & 11.58 \\
 & 0.25 & 6.60 & 6.88 & 9.78 & 10.47 & 10.76 & 11.73 \\
 & 0.50 & 6.75 & 6.93 & 9.93 & 10.59 & 10.89 & 11.82 \\
 & 0.75 & 6.55 & 6.85 & 9.72 & 10.50 & 10.70 & 11.66 \\
 & 1.00 & 6.12 & 6.33 & 9.18 & 10.10 & 10.38 & 11.31 \\
\bottomrule
\end{tabular}%
}
\endgroup
\end{table*}

\begin{table*}[ht!]
\centering
\caption{Invocation rates (\%) of different models for different choices of \(\lambda\) values.}
\label{tab:lambda-ablation-call-percentages}
\resizebox{\textwidth}{!}{%
\begin{tabular}{@{}llccccccccc@{}}
\toprule
Benchmark & $\lambda$ / Models &
\makecell{DeepSeek-R1-\\Distill-Qwen-7B} &
\makecell{Llama-3.1-\\8B-Instruct} &
Qwen3-8B &
Qwen3-14B &
\makecell{Devstral-\\Small-2505} &
gpt-5-mini &
\makecell{DeepSeek-R1-\\Distill-Qwen-32B} &
gpt-5.2 &
\makecell{gpt-5.2\\Course Alteration} \\
\midrule

\multirow{5}{*}{Llama-3-8B Attention Layer}
 & 0.00 & 16.2\% & 18.7\% & 10.0\% & 7.4\% & 1.3\% & 5.8\% & 4.8\% & 23.1\% & 12.7\% \\
 & 0.25 & 13.2\% & 6.5\% & 28.0\% & 7.2\% & 3.6\% & 7.2\% & 1.0\% & 19.8\% & 13.5\% \\
 & 0.50 & 11.6\% & 7.6\% & 34.1\% & 5.1\% & 1.0\% & 5.6\% & 2.5\% & 18.3\% & 14.3\% \\
 & 0.75 & 15.4\% & 22.4\% & 16.8\% & 7.8\% & 1.0\% & 5.3\% & 1.8\% & 14.4\% & 15.1\% \\
 & 1.00 & 14.4\% & 9.2\% & 32.1\% & 3.8\% & 4.5\% & 6.1\% & 1.3\% & 12.8\% & 16.0\% \\
\midrule

\multirow{5}{*}{DeepSeek-R1 MoE Layer}
 & 0.00 & 10.3\% & 20.9\% & 7.3\% & 11.1\% & 3.0\% & 21.7\% & 1.9\% & 12.4\% & 11.4\% \\
 & 0.25 & 12.6\% & 16.0\% & 10.7\% & 11.7\% & 4.0\% & 20.2\% & 1.7\% & 10.9\% & 11.9\% \\
 & 0.50 & 12.4\% & 23.5\% & 8.9\% & 1.8\% & 2.7\% & 27.3\% & 1.3\% & 9.6\% & 12.4\% \\
 & 0.75 & 11.9\% & 25.7\% & 10.2\% & 1.6\% & 4.2\% & 22.0\% & 2.7\% & 9.0\% & 12.8\% \\
 & 1.00 & 14.8\% & 25.7\% & 11.6\% & 3.9\% & 4.5\% & 16.4\% & 1.5\% & 9.4\% & 12.4\% \\
\midrule

\multirow{5}{*}{FLUX Attention Layer}
 & 0.00 & 12.6\% & 13.8\% & 22.3\% & 11.9\% & 3.9\% & 3.3\% & 8.2\% & 11.5\% & 12.4\% \\
 & 0.25 & 30.6\% & 14.5\% & 16.5\% & 2.7\% & 3.5\% & 2.5\% & 4.9\% & 12.0\% & 12.8\% \\
 & 0.50 & 13.7\% & 14.9\% & 36.5\% & 2.3\% & 3.4\% & 1.7\% & 4.9\% & 9.2\% & 13.3\% \\
 & 0.75 & 5.9\% & 28.3\% & 20.8\% & 10.7\% & 5.1\% & 3.0\% & 3.7\% & 8.5\% & 13.9\% \\
 & 1.00 & 21.3\% & 16.0\% & 24.8\% & 7.5\% & 2.1\% & 0.9\% & 5.1\% & 7.4\% & 14.2\% \\
\midrule

\multirow{5}{*}{FLUX Convolution Layer}
 & 0.00 & 11.6\% & 21.2\% & 25.2\% & 10.8\% & 4.1\% & 2.1\% & 0.8\% & 12.8\% & 11.4\% \\
 & 0.25 & 5.9\% & 29.9\% & 25.0\% & 7.3\% & 5.1\% & 4.9\% & 0.4\% & 8.8\% & 12.5\% \\
 & 0.50 & 6.2\% & 3.7\% & 54.5\% & 11.2\% & 4.3\% & 0.5\% & 0.3\% & 5.7\% & 13.5\% \\
 & 0.75 & 23.9\% & 27.8\% & 12.7\% & 10.6\% & 5.2\% & 0.6\% & 1.0\% & 0.4\% & 14.1\% \\
 & 1.00 & 23.1\% & 21.3\% & 22.6\% & 7.8\% & 2.4\% & 3.0\% & 1.4\% & 3.8\% & 14.6\% \\
\midrule

\multirow{5}{*}{Llama-4-Scout MLP Layer}
 & 0.00 & 16.0\% & 30.4\% & 12.2\% & 6.5\% & 1.0\% & 5.6\% & 2.0\% & 14.2\% & 12.0\% \\
 & 0.25 & 17.2\% & 31.8\% & 13.9\% & 2.0\% & 1.5\% & 7.5\% & 1.4\% & 12.3\% & 12.4\% \\
 & 0.50 & 19.0\% & 7.3\% & 41.6\% & 3.1\% & 0.8\% & 4.2\% & 0.9\% & 10.6\% & 12.6\% \\
 & 0.75 & 20.5\% & 24.7\% & 25.7\% & 2.7\% & 0.7\% & 3.4\% & 0.6\% & 8.9\% & 12.9\% \\
 & 1.00 & 19.4\% & 31.6\% & 18.0\% & 4.9\% & 1.3\% & 2.7\% & 2.4\% & 6.4\% & 13.2\% \\
\bottomrule
\end{tabular}%
}
\end{table*}
\clearpage
\section{Course Alteration Ablations}
\label{sec:course-alteration-ablation}
Across the five benchmarks, triggering course alteration after every small model regression achieves the highest speedup, whereas disabling course alteration yields the lowest speedup. \colt instead triggers course alteration after every two small model regressions, providing the best speedup-cost tradeoff. As shown in Table \ref{tab:course-alteration-ablation-call-percentages} and Table \ref{tab:litecoop_improvement_course_alteration}, after 1000 searched samples, \colteight lowers the largest model course alteration rate from 29.9\% to 13.2\%, while reducing compilation time by 1.38\(\times\) and API cost by 1.58\(\times\) (geometric means over the five benchmarks). This reduced course alteration rate incurs only a 1.3\% mean relative final-speedup gap per kernel, as shown in Table \ref{tab:course-alteration-ablation-speedups} (1.5 points of geometric-mean gain over \rc; \tref{tab:ablation}).

\begin{table*}[ht]
\centering
\caption{Speedup over unoptimized code across varying numbers of samples for different course alteration settings.}
\label{tab:course-alteration-ablation-speedups}
\resizebox{\textwidth}{!}{%
\begin{tabular}{@{}llcccccc@{}}
\toprule
 Benchmark & \makecell{Course Alteration / Number of Samples} & 50 & 100 & 250 & 500 & 750 & 1000 \\
\midrule

\multirow{3}{*}{Llama-3-8B Attention Layer}
 & No Course Alteration & 6.91 & 10.02 & 10.19 & 11.78 & 12.01 & 14.05 \\
 & Every 1 Small Model Regression & 7.60 & 10.68 & 11.16 & 12.88 & 13.07 & 15.26 \\
 & Every 2 Small Model Regressions & 7.56 & 10.61 & 10.99 & 12.65 & 12.81 & 14.98 \\
\midrule

\multirow{3}{*}{DeepSeek-R1 MoE Layer}
 & No Course Alteration & 6.92 & 8.57 & 10.88 & 11.85 & 11.91 & 12.91 \\
 & Every 1 Small Model Regression & 7.56 & 9.29 & 12.09 & 12.31 & 12.40 & 14.24 \\
 & Every 2 Small Model Regressions & 6.99 & 8.66 & 11.84 & 12.20 & 12.28 & 13.87 \\
\midrule

\multirow{3}{*}{FLUX Attention Layer}
 & No Course Alteration & 5.07 & 5.58 & 6.55 & 6.94 & 7.02 & 7.61 \\
 & Every 1 Small Model Regression & 5.50 & 5.74 & 6.82 & 7.23 & 7.32 & 8.18 \\
 & Every 2 Small Model Regressions & 5.47 & 5.69 & 6.72 & 7.18 & 7.24 & 8.12 \\
\midrule

\multirow{3}{*}{FLUX Convolution Layer}
 & No Course Alteration & 2.84 & 3.07 & 4.11 & 4.43 & 4.78 & 5.33 \\
 & Every 1 Small Model Regression & 2.91 & 3.87 & 4.25 & 4.61 & 5.10 & 5.53 \\
 & Every 2 Small Model Regressions & 2.88 & 3.83 & 4.19 & 4.53 & 5.01 & 5.51 \\
\midrule

\multirow{3}{*}{Llama-4-Scout MLP Layer}
 & No Course Alteration & 5.57 & 6.18 & 8.53 & 9.54 & 10.55 & 11.36 \\
 & Every 1 Small Model Regression & 6.87 & 7.08 & 10.02 & 10.78 & 10.98 & 11.92 \\
 & Every 2 Small Model Regressions & 6.75 & 6.93 & 9.93 & 10.59 & 10.89 & 11.82 \\
\bottomrule
\end{tabular}%
}
\end{table*}

\begin{table*}[ht]
\centering
\caption{Invocation rates (\%) of different models for different course alteration settings.}
\label{tab:course-alteration-ablation-call-percentages}
\vspace{-0.1cm}
\resizebox{\textwidth}{!}{%
\begin{tabular}{@{}llccccccccc@{}}
\toprule

Benchmark & \makecell{Course Alteration / Models} &
\makecell{DeepSeek-R1-\\Distill-Qwen-7B} &
\makecell{Llama-3.1-\\8B-Instruct} &
Qwen3-8B &
Qwen3-14B &
\makecell{Devstral-\\Small-2505} &
gpt-5-mini &
\makecell{DeepSeek-R1-\\Distill-Qwen-32B} &
gpt-5.2 &
\makecell{gpt-5.2\\Course Alteration} \\
\midrule

\multirow{3}{*}{Llama-3-8B Attention Layer}
 & No Course Alteration & 13.8\% & 17.5\% & 30.2\% & 5.3\% & 1.4\% & 7.8\% & 2.6\% & 21.4\% & 0.0\% \\
 & Every 1 Small Model Reg. & 13.8\% & 12.5\% & 14.8\% & 4.0\% & 2.2\% & 8.7\% & 1.5\% & 11.7\% & 30.7\% \\
 & Every 2 Small Model Reg. & 11.6\% & 7.6\% & 34.1\% & 5.1\% & 1.0\% & 5.6\% & 2.5\% & 18.3\% & 14.3\% \\
\midrule

\multirow{3}{*}{DeepSeek-R1 MoE Layer}
 & No Course Alteration & 18.9\% & 23.7\% & 10.5\% & 1.1\% & 2.7\% & 30.9\% & 1.4\% & 10.8\% & 0.0\% \\
 & Every 1 Small Model Reg. & 11.7\% & 15.8\% & 6.5\% & 3.4\% & 2.9\% & 21.2\% & 1.2\% & 8.7\% & 28.4\% \\
 & Every 2 Small Model Reg. & 12.4\% & 23.5\% & 8.9\% & 1.8\% & 2.7\% & 27.3\% & 1.3\% & 9.6\% & 12.4\% \\
\midrule

\multirow{3}{*}{FLUX Attention Layer}
 & No Course Alteration & 15.9\% & 13.4\% & 22.6\% & 11.6\% & 8.1\% & 11.7\% & 1.3\% & 15.4\% & 0.0\% \\
 & Every 1 Small Model Reg. & 17.6\% & 11.1\% & 8.6\% & 4.9\% & 2.6\% & 16.3\% & 0.8\% & 9.4\% & 28.7\% \\
 & Every 2 Small Model Reg. & 13.7\% & 14.9\% & 36.5\% & 2.3\% & 3.4\% & 1.7\% & 4.9\% & 9.2\% & 13.3\% \\
\midrule

\multirow{3}{*}{FLUX Convolution Layer}
 & No Course Alteration & 9.8\% & 25.2\% & 32.7\% & 13.4\% & 4.8\% & 5.7\% & 0.6\% & 7.8\% & 0.0\% \\
 & Every 1 Small Model Reg. & 12.2\% & 17.0\% & 21.1\% & 9.0\% & 3.4\% & 0.8\% & 1.0\% & 5.7\% & 29.8\% \\
 & Every 2 Small Model Reg. & 6.2\% & 3.7\% & 54.5\% & 11.2\% & 4.3\% & 0.5\% & 0.3\% & 5.7\% & 13.5\% \\
\midrule

\multirow{3}{*}{Llama-4-Scout MLP Layer}
 & No Course Alteration & 21.1\% & 8.2\% & 34.9\% & 3.4\% & 9.2\% & 7.7\% & 0.9\% & 14.6\% & 0.0\% \\
 & Every 1 Small Model Reg. & 15.1\% & 19.5\% & 14.3\% & 2.5\% & 4.0\% & 3.3\% & 2.5\% & 7.2\% & 31.7\% \\
 & Every 2 Small Model Reg. & 19.0\% & 7.3\% & 41.6\% & 3.1\% & 0.8\% & 4.2\% & 0.9\% & 10.6\% & 12.6\% \\
\bottomrule
\end{tabular}%
}
\end{table*}

\begin{table*}[h]
    \centering
    \caption{Compilation time and API cost reduction of performing course alteration every two small model regression against performing course alteration every one small model regression.}
    \label{tab:litecoop_improvement_course_alteration}
    \renewcommand{\arraystretch}{1.08}
    \setlength{\tabcolsep}{3.4pt}
    \small
    
    \begin{tabular}{lccccc}
        \toprule
         &
        \makecell{Llama-3-8B \\ Attention Layer} &
        \makecell{DeepSeek-R1 \\ MoE Layer} &
        \makecell{FLUX \\ Attention Layer} &
        \makecell{FLUX \\ Convolution Layer} &
        \makecell{Llama-4-Scout \\ MLP Layer} \\
    
        \midrule
            
        Comp. Time $\downarrow$ ($\times$)
        & 1.29 & 1.34 & 1.53 & 1.37 & 1.38 \\
        
        API Cost $\downarrow$ ($\times$)
        & 1.30 & 1.57 & 1.69 & 1.80 & 1.58 \\
        
        \bottomrule
    \end{tabular}
\end{table*}

\clearpage
\section{LLM Selection Ablations}
\label{sec:model-selection-ablation}

\colt's gains are not due to LLM diversity alone. To isolate the role of endogenous LLM selection, we compare \colteight against two ablations that use the same eight LLM pool but replace \colt's state-dependent routing with random or round-robin next model selection. Unlike these static policies, \colt makes the next model choice part of the joint action: in each expansion, the active LLM considers the current program context, search progress and per model statistics, and proposes both a compiler transformation and the LLM that should expand the resulting child. As shown in Table~\ref{tab:model-selection-ablation-speedups}, \colteight achieves the best speedup at every reported number of searched samples on every benchmark. For example, \colteight achieves 16.8\% and 15.6\% higher final speedup compared with random and round-robin LLM selection respectively on the \llamabench.  The same trend holds for tuning cost: Table~\ref{tab:litecoop_improvement_model_selection} shows that \colteight reduces compilation time by 1.37$\times$/1.38$\times$ and API cost by 1.27$\times$/1.26$\times$ (geometric means over the five benchmarks) relative to random/round-robin selection. These results show that random and round-robin LLM selection fail to choose the LLMs with the best cost-quality tradeoffs. Therefore, on CPU, the benefit of \colt over these static policies comes from treating model choice as part of the joint MCTS action, rather than simply exposing the compiler to a heterogeneous set of LLMs.

\begin{table*}[ht]
\vspace{-0.2cm}
\centering
\caption{Speedup over unoptimized code across varying numbers of samples for different LLM selection settings.}
\label{tab:model-selection-ablation-speedups}
\vspace{-0.1cm}
\resizebox{\textwidth}{!}{%
\begin{tabular}{@{}llcccccc@{}}
\toprule
Benchmark 
& \makecell{LLM Selection / Number of Samples} 
& 50 & 100 & 250 & 500 & 750 & 1000 \\
\midrule

\multirow{3}{*}{Llama-3-8B Attention Layer}
 & \colteight & 7.56 & 10.61 & 10.99 & 12.65 & 12.81 & 14.98 \\
 & Random & 6.20 & 8.55 & 10.05 & 10.60 & 11.10 & 12.82 \\
 & Round-Robin & 6.30 & 8.70 & 10.18 & 10.55 & 11.25 & 12.96 \\
\midrule

\multirow{3}{*}{DeepSeek-R1 MoE Layer}
 & \colteight & 6.99 & 8.66 & 11.84 & 12.20 & 12.28 & 13.87 \\
 & Random & 5.28 & 7.62 & 10.72 & 11.90 & 11.98 & 12.54 \\
 & Round-Robin & 5.06 & 6.60 & 10.89 & 11.96 & 11.93 & 12.69 \\
\midrule

\multirow{3}{*}{FLUX Attention Layer}
 & \colteight & 5.47 & 5.69 & 6.72 & 7.18 & 7.24 & 8.12 \\
 & Random & 4.61 & 5.53 & 6.43 & 6.80 & 6.84 & 7.37 \\
 & Round-Robin & 3.74 & 5.01 & 6.36 & 6.85 & 6.88 & 7.42 \\
\midrule

\multirow{3}{*}{FLUX Convolution Layer}
 & \colteight & 2.88 & 3.83 & 4.19 & 4.53 & 5.01 & 5.51 \\
 & Random & 2.24 & 2.73 & 3.35 & 3.91 & 4.65 & 4.93 \\
 & Round-Robin & 2.32 & 2.53 & 3.08 & 3.78 & 4.63 & 4.82 \\
\midrule

\multirow{3}{*}{Llama-4-Scout MLP Layer}
 & \colteight & 6.75 & 6.93 & 9.93 & 10.59 & 10.89 & 11.82 \\
 & Random & 3.76 & 4.81 & 7.60 & 9.36 & 10.08 & 10.83 \\
 & Round-Robin & 3.92 & 4.85 & 7.62 & 9.28 & 9.90 & 10.80 \\
\bottomrule
\end{tabular}%
}
\end{table*}

\vspace{-0.3cm}
\begin{table*}[ht]
\centering
\caption{Invocation rates (\%) of different models for different LLM selection settings.}
\label{tab:model-selection-ablation-call-percentages}
\vspace{-0.1cm}
\resizebox{\textwidth}{!}{%
\begin{tabular}{@{}llccccccccc@{}}
\toprule
Benchmark
& \makecell{LLM Selection / Models}
& \makecell{DeepSeek-R1-\\Distill-Qwen-7B}
& \makecell{Llama-3.1-\\8B-Instruct}
& Qwen3-8B
& Qwen3-14B
& \makecell{Devstral-\\Small-2505}
& gpt-5-mini
& \makecell{DeepSeek-R1-\\Distill-Qwen-32B}
& gpt-5.2
& \makecell{gpt-5.2\\Course Alteration} \\
\midrule

\multirow{3}{*}{Llama-3-8B Attention Layer}
 & \colteight & 11.6\% & 7.6\% & 34.1\% & 5.1\% & 1.0\% & 5.6\% & 2.5\% & 18.3\% & 14.3\% \\
 & Random & 10.5\% & 10.4\% & 11.5\% & 9.9\% & 11.2\% & 10.7\% & 10.6\% & 10.1\% & 15.0\% \\
 & Round-Robin & 10.7\% & 10.7\% & 10.7\% & 10.7\% & 10.7\% & 10.7\% & 10.7\% & 10.7\% & 14.4\% \\
\midrule

\multirow{3}{*}{DeepSeek-R1 MoE Layer}
 & \colteight & 12.4\% & 23.5\% & 8.9\% & 1.8\% & 2.7\% & 27.3\% & 1.3\% & 9.6\% & 12.4\% \\
 & Random & 11.3\% & 10.2\% & 11.5\% & 10.0\% & 10.7\% & 11.0\% & 10.3\% & 11.1\% & 13.9\% \\
 & Round-Robin & 10.9\% & 10.9\% & 10.9\% & 10.9\% & 10.9\% & 10.9\% & 10.9\% & 10.9\% & 12.9\% \\
\midrule

\multirow{3}{*}{FLUX Attention Layer}
 & \colteight & 13.7\% & 14.9\% & 36.5\% & 2.3\% & 3.4\% & 1.7\% & 4.9\% & 9.2\% & 13.3\% \\
 & Random & 11.7\% & 12.9\% & 12.7\% & 10.7\% & 10.2\% & 8.8\% & 10.7\% & 8.2\% & 14.1\% \\
 & Round-Robin & 10.8\% & 10.8\% & 10.8\% & 10.8\% & 10.8\% & 10.8\% & 10.8\% & 10.8\% & 13.5\% \\
\midrule

\multirow{3}{*}{FLUX Convolution Layer}
 & \colteight & 6.2\% & 3.7\% & 54.5\% & 11.2\% & 4.3\% & 0.5\% & 0.3\% & 5.7\% & 13.5\% \\
 & Random & 10.7\% & 11.0\% & 11.1\% & 10.1\% & 10.5\% & 10.9\% & 10.6\% & 11.4\% & 13.7\% \\
 & Round-Robin & 10.7\% & 10.7\% & 10.7\% & 10.7\% & 10.7\% & 10.7\% & 10.7\% & 10.7\% & 14.1\% \\
\midrule

\multirow{3}{*}{Llama-4-Scout MLP Layer}
 & \colteight & 19.0\% & 7.3\% & 41.6\% & 3.1\% & 0.8\% & 4.2\% & 0.9\% & 10.6\% & 12.6\% \\
 & Random & 11.5\% & 10.8\% & 11.0\% & 10.1\% & 10.7\% & 11.5\% & 10.2\% & 11.1\% & 13.2\% \\
 & Round-Robin & 10.8\% & 10.8\% & 10.8\% & 10.8\% & 10.8\% & 10.8\% & 10.8\% & 10.8\% & 13.6\% \\
\bottomrule
\end{tabular}%
}
\end{table*}

\vspace{-0.3cm}
\begin{table*}[h]
    \centering
    \caption{Compilation time and API cost reduction of \colteight against random / round-robin LLM selection settings.}
    \label{tab:litecoop_improvement_model_selection}
    \vspace{-0.1cm}
    \renewcommand{\arraystretch}{1.08}
    \setlength{\tabcolsep}{3.4pt}
    \small
    
    \begin{tabular}{lccccc}
        \toprule
         &
        \makecell{Llama-3-8B \\ Attention Layer} &
        \makecell{DeepSeek-R1 \\ MoE Layer} &
        \makecell{FLUX \\ Attention Layer} &
        \makecell{FLUX \\ Convolution Layer} &
        \makecell{Llama-4-Scout \\ MLP Layer} \\
    
        \midrule
            
        Comp. Time $\downarrow$ ($\times$)
         & 1.17 / 1.18 & 1.21 / 1.24 & 1.39 / 1.45 & 1.49 / 1.44 & 1.64 / 1.62 \\
        
        API Cost $\downarrow$ ($\times$)
        & 1.05 / 1.03 & 1.13 / 1.15 & 1.29 / 1.31 & 1.55 / 1.53 & 1.38 / 1.36 \\
        
        \bottomrule
    \end{tabular}
\end{table*}
\clearpage
\section{Number of Model Invocations}

\begin{table*}[ht]
\centering
\caption{Model call counts for 2-, 4-, and 8-LLM configurations of \colt across five benchmarks on GPU when \gptfivetwo is the largest model.}
\label{tab:model-call-counts-3}
\renewcommand{\arraystretch}{1.1}
\resizebox{\textwidth}{!}{%
\begin{tabular}{@{}llccccccccc@{}}
\toprule

\multicolumn{2}{c}{} &
\multicolumn{8}{c}{Number of Regular Model Calls} &
\multicolumn{1}{c}{Number of Course Alterations} \\
\cmidrule(lr){3-10}\cmidrule(lr){11-11}
Benchmark & \makecell{\colt \\ Combination} &
\makecell{DeepSeek-\\R1-Distill-\\Qwen-7B} &
\makecell{Llama-3.1-\\8B-\\Instruct} &
\makecell{Qwen3-\\8B} &
\makecell{Qwen3-\\14B} &
\makecell{Devstral-\\Small-\\2505} &
\makecell{gpt-5-\\mini} &
\makecell{DeepSeek-\\R1-Distill-\\Qwen-32B} &
\gptfivetwo &
\gptfivetwo \\
\midrule

\multirow{3}{*}{\llamathreeb Attention Layer}
  & \colteight & 183 & 21  & 411 & 94  & 43  & 60  & 79  & 109 & 158 \\
  & \coltfour  & --- & 358 & --- & --- & --- & 377 & 99  & 166 & 142 \\
  & \colttwo   & --- & --- & --- & --- & --- & 588 & --- & 412 & 103 \\
\midrule

\multirow{3}{*}{\deepseekrone MoE Layer}
  & \colteight & 257 & 129 & 86  & 417 & 0   & 14  & 18  & 79  & 162 \\
  & \coltfour  & --- & 198 & --- & --- & --- & 576 & 30  & 196 & 146 \\
  & \colttwo   & --- & --- & --- & --- & --- & 764 & --- & 236 & 120 \\
\midrule

\multirow{3}{*}{\flux Attention Layer}
  & \colteight & 146 & 47  & 465 & 216 & 14  & 7   & 26  & 79  & 155 \\
  & \coltfour  & --- & 419 & --- & --- & --- & 307 & 92  & 182 & 140 \\
  & \colttwo   & --- & --- & --- & --- & --- & 584 & --- & 416 & 88 \\
\midrule

\multirow{3}{*}{\flux Convolution Layer}
  & \colteight & 316 & 196 & 130 & 72  & 3   & 13  & 112 & 158 & 124 \\
  & \coltfour  & --- & 350 & --- & --- & --- & 392 & 78  & 180 & 137 \\
  & \colttwo   & --- & --- & --- & --- & --- & 813 & --- & 187 & 142 \\
\midrule

\multirow{3}{*}{\llamafourscout MLP Layer}
  & \colteight & 565 & 35  & 168 & 54  & 6   & 8   & 25  & 139 & 167 \\
  & \coltfour  & --- & 338 & --- & --- & --- & 246 & 223 & 193 & 127 \\
  & \colttwo   & --- & --- & --- & --- & --- & 788 & --- & 212 & 139 \\
\bottomrule
\end{tabular}%
}
\end{table*}

\begin{table*}[ht]
\centering
\caption{Model call counts for 2-, 4-, and 8-LLM configurations of \colt across five benchmarks on CPU when \gptfivetwo is the largest model.}
\label{tab:model-call-counts}
\renewcommand{\arraystretch}{1.1}
\resizebox{\textwidth}{!}{%
\begin{tabular}{@{}llccccccccc@{}}
\toprule

\multicolumn{2}{c}{} &
\multicolumn{8}{c}{Number of Regular Model Calls} &
\multicolumn{1}{c}{Number of Course Alterations} \\
\cmidrule(lr){3-10}\cmidrule(lr){11-11}
Layer & Experiment &
\makecell{DeepSeek-\\R1-Distill-\\Qwen-7B} &
\makecell{Llama-3.1-\\8B-\\Instruct} &
\makecell{Qwen3-\\8B} &
\makecell{Qwen3-\\14B} &
\makecell{Devstral-\\Small-\\2505} &
\makecell{gpt-5-\\mini} &
\makecell{DeepSeek-\\R1-Distill-\\Qwen-32B} &
\gptfivetwo &
\gptfivetwo \\
\midrule

\multirow{3}{*}{\llamathreeb Attention Layer}
  & \colteight & 135 & 89  & 398 & 59  & 12  & 65  & 29  & 213 & 167 \\
  & \coltfour  & --- & 163 & --- & --- & --- & 229 & 365 & 243 & 128 \\
  & \colttwo   & --- & --- & --- & --- & --- & 776 & --- & 224 & 110 \\
\midrule

\multirow{3}{*}{\deepseekrone MoE Layer}
  & \colteight & 142 & 268 & 102 & 21  & 31  & 312 & 15  & 109 & 141 \\
  & \coltfour  & --- & 346 & --- & --- & --- & 181 & 301 & 172 & 137 \\
  & \colttwo   & --- & --- & --- & --- & --- & 743 & --- & 257 & 125 \\
\midrule

\multirow{3}{*}{\flux Attention Layer}
  & \colteight & 158 & 172 & 421 & 27  & 39  & 20  & 57  & 106 & 154 \\
  & \coltfour  & --- & 77  & --- & --- & --- & 536 & 208 & 179 & 139 \\
  & \colttwo   & --- & --- & --- & --- & --- & 727 & --- & 273 & 114 \\
\midrule

\multirow{3}{*}{\flux Convolution Layer}
  & \colteight & 72  & 43  & 630 & 129 & 50  & 6   & 4   & 66  & 156 \\
  & \coltfour  & --- & 217 & --- & --- & --- & 539 & 57  & 187 & 143 \\
  & \colttwo   & --- & --- & --- & --- & --- & 686 & --- & 314 & 122 \\
\midrule

\multirow{3}{*}{\llamafourscout MLP Layer}
  & \colteight & 217 & 84  & 476 & 35  & 9   & 48  & 10  & 121 & 144 \\
  & \coltfour  & --- & 55  & --- & --- & --- & 587 & 128 & 230 & 136 \\
  & \colttwo   & --- & --- & --- & --- & --- & 698 & --- & 302 & 118 \\
\bottomrule
\end{tabular}%
}
\end{table*}

\begin{table*}[ht]
\centering
\caption{Model call counts for 2-, 4-, and 8-LLM configurations of \colt across five benchmarks on CPU when \llamathreeseventyb is the largest model.}
\label{tab:model-call-counts-2}
\renewcommand{\arraystretch}{1.1}
\resizebox{\textwidth}{!}{%
\begin{tabular}{@{}llccccccccc@{}}
\toprule

\multicolumn{2}{c}{} &
\multicolumn{8}{c}{Number of Regular Model Calls} &
\multicolumn{1}{c}{Number of Course Alterations} \\
\cmidrule(lr){3-10}\cmidrule(lr){11-11}
Layer & Experiment &
\makecell{DeepSeek-\\R1-Distill-\\Qwen-7B} &
\makecell{Llama-3.1-\\8B-\\Instruct} &
\makecell{Qwen3-\\8B} &
\makecell{Qwen3-\\14B} &
\makecell{Devstral-\\Small-\\2505} &
\makecell{gpt-5-\\mini} &
\makecell{DeepSeek-\\R1-Distill-\\Qwen-32B} &
\makecell{Llama-3.3-\\70B-\\Instruct} &
\makecell{Llama-3.3-\\70B-\\Instruct} \\
\midrule

\multirow{3}{*}{\llamathreeb Attention Layer}
  & \colteight & 51  & 54  & 372 & 108 & 88  & 92  & 6   & 229 & 144 \\
  & \coltfour  & --- & 214 & --- & --- & --- & 329 & 122 & 335 & 94 \\
  & \colttwo   & --- & --- & --- & --- & --- & 495 & --- & 505 & 87 \\
\midrule

\multirow{3}{*}{\deepseekrone MoE Layer}
  & \colteight & 98  & 88  & 213 & 368 & 20  & 5   & 2   & 206 & 139 \\
  & \coltfour  & --- & 416 & --- & --- & --- & 159 & 69  & 356 & 122 \\
  & \colttwo   & --- & --- & --- & --- & --- & 539 & --- & 461 & 91 \\
\midrule

\multirow{3}{*}{\flux Attention Layer}
  & \colteight & 116 & 298 & 114 & 93  & 37  & 8   & 6   & 328 & 125 \\
  & \coltfour  & --- & 230 & --- & --- & --- & 379 & 49  & 342 & 119 \\
  & \colttwo   & --- & --- & --- & --- & --- & 549 & --- & 451 & 93 \\
\midrule

\multirow{3}{*}{\flux Convolution Layer}
  & \colteight & 140 & 277 & 153 & 122 & 52  & 63  & 39  & 154 & 130 \\
  & \coltfour  & --- & 198 & --- & --- & --- & 284 & 121 & 397 & 96 \\
  & \colttwo   & --- & --- & --- & --- & --- & 516 & --- & 484 & 82 \\
\midrule

\multirow{3}{*}{\llamafourscout MLP Layer}
  & \colteight & 103 & 172 & 400 & 176 & 2   & 2   & 4   & 141 & 153 \\
  & \coltfour  & --- & 224 & --- & --- & --- & 399 & 109 & 268 & 129 \\
  & \colttwo   & --- & --- & --- & --- & --- & 429 & --- & 571 & 75 \\
\bottomrule
\end{tabular}%
}
\end{table*}

\newpage


\end{document}